\RequirePackage{fix-cm}
\documentclass[smallextended]{svjour3}       
\smartqed  
\usepackage{graphicx}

\usepackage{subfig}
\usepackage{amsfonts}
\usepackage{cancel}
\usepackage{xcolor}

\usepackage{natbib}
\setcitestyle{authoryear,open={(},close={)}} 
\setcitestyle{aysep={}} 

\usepackage[]{algorithm2e}
\makeatletter
\renewcommand{\@algocf@capt@plain}{above}
\makeatother
\usepackage{algorithmic}
\newif\ifworkinprogress
\workinprogresstrue
\ifworkinprogress
  \newcommand{\se}[1]{\textcolor{blue}{\textbf{[Samaneh] #1}}}
  \newcommand{\crn}[1]{\textcolor{magenta}{\textbf{[Chitta] #1}}}
  \newcommand{\kp}[1]{\textcolor{brown}{\textbf{[Kamran] #1}}}
\else
  \newcommand{\se}[1]{}
  \newcommand{\crn}[1]{}
  \newcommand{\kp}[1]{}
\fi

\usepackage{amsmath}  

\usepackage{listings}
\usepackage{xcolor}

\definecolor{codegreen}{rgb}{0,0.6,0}
\definecolor{codegray}{rgb}{0.5,0.5,0.5}
\definecolor{codepurple}{rgb}{0.58,0,0.82}
\definecolor{backcolour}{rgb}{0.95,0.95,0.92}

\lstdefinestyle{mystyle}{
    backgroundcolor=\color{backcolour},   
    commentstyle=\color{codegreen},
    keywordstyle=\color{magenta},
    numberstyle=\tiny\color{codegray},
    stringstyle=\color{codepurple},
    basicstyle=\ttfamily\footnotesize,
    breakatwhitespace=false,         
    breaklines=true,                 
    captionpos=b,                    
    keepspaces=true,                 
    numbers=left,                    
    numbersep=5pt,                  
    showspaces=false,                
    showstringspaces=false,
    showtabs=false,                  
    tabsize=2
}

\usepackage{cases}

\newcommand{\indep}{\perp \!\!\! \perp}

\usepackage{hyperref}

\begin{document}

\title{Sequence Graph Transform (SGT): \\
			A Feature Embedding Function for Sequence Data Mining 
}

\titlerunning{Sequence Graph Transform}        

\author{Chitta Ranjan         \and
        Samaneh Ebrahimi  	  \and
        Kamran Paynabar
}


\institute{Chitta Ranjan \at
          H. Milton Stewart School of Industrial and Systems Engineering,\\
          Georgia Institute of Technology, Atlanta, GA, USA \\
              \email{nk.chitta.ranjan@gatech.edu}
           \and
           Samaneh Ebrahimi \at
           H. Milton Stewart School of Industrial and Systems Engineering,
           \\
           Georgia Institute of Technology, Atlanta, GA, USA \\
              \email{samaneh.ebrahimi@gatech.edu}
           \and
           Kamran Paynabar \at
           H. Milton Stewart School of Industrial and Systems Engineering, \\
           Georgia Institute of Technology, Atlanta, GA, USA
           \\
              \email{kamran.paynabar@isye.gatech.edu}
}

\date{Received: date / Accepted: date}

\maketitle

\begin{abstract}
Sequence feature embedding is a challenging task due to the unstructuredness of
sequences---arbitrary strings of arbitrary length. Existing methods are efficient
in extracting short-term dependencies but typically suffer from computation issues
for the long-term. Sequence Graph Transform (SGT), a feature embedding
function, that can extract a varying amount of short- to long-term dependencies without
increasing the computation is proposed. SGT's properties are analytically proved for interpretation under normal and uniform distribution assumptions. SGT features yield significantly superior results 
in sequence clustering and classification with higher accuracy and lower computation as compared to the existing methods, including the state-of-the-art sequence/string Kernels and LSTM.


\end{abstract}

\section{Introduction}
\label{sec:Introduction}

A sequence is an ordered series of discrete items, where each
item can be a bucket of elements. For example, $\{\langle B \rangle
\langle AAB \rangle \langle CC \rangle \langle A \rangle \}$.
A commonly found specific case of sequences is when each item has only one
element, e.g., $\{\langle \mathtt{B} \rangle \langle \mathtt{A} \rangle \langle \mathtt{A} \rangle \langle \mathtt{B} \rangle \langle \mathtt{C} \rangle \langle \mathtt{C} \rangle \langle \mathtt{A} \rangle \}$ or simply put $\mathtt{BAABCCA}$. In this paper, 
a work has been done on this class of sequences (a.k.a strings) that we define
as a series of discrete \emph{symbols} sequentially tied together in a certain order. A symbol can be an event, or a value. Such sequences are found in
processes where only one discrete event can occur at one time, such as
clickstream, music listening history, weblog, patient movements, and protein
sequences.

Sequence data is omnipresent which has led to the development of
various sequence mining methods. Sequence mining research can be broadly
divided into: a) frequent pattern or subsequence mining \citep{aggarwal2014frequent}, 
b) motifs detection \citep{sandve2006survey}, c) alignment \citep{li2010survey}, 
d) datastream modeling \citep{silva2013data}, 
and e) feature embedding \citep{kumar2012pattern}. Among them,
 feature embedding is particularly important because it provides a machine-interpretable representation for the sequences. They can be used directly for (dis)similarity or ``distance'' computation between sequences or other machine learning models. 
A similar approach \textit{word2vec} is popular in text mining for converting text into vector embeddings. This enables building sequence classification 
and clustering models, which have immense applications 
across the online industry, Bioinformatics, and healthcare.


Feature embedding is, however, challenging because a) sequences
are arbitrary strings of arbitrary lengths, and b) long-term dependencies
(of sequence elements) are difficult to capture. A long-term dependency here means the effect of distant elements in a sequence on each other.

\textit{N}-gram methods (also known as \textit{k}-mers) are commonly used for feature
representation. Several sequence kernels are developed on top of the \textit{n}-grams features.
Moreover, generative parametric models, such as \textit{n}-order Markov and HMM models
have been developed for sequences in which sequence features are 
represented by the transition and emission probability matrices.

However, in addition to other limitations (discussed in \S\ref{subsec:related-work}),
most of the existing methods either limit themselves by extracting only
short-term patterns or suffer from increasing computation
upon extracting the long-term patterns.

Additionally, accurately comparing sequences of different lengths is
a non-trivial problem. Traditional methods often lead to \textit{false positives}.
A false positive here implies incorrectly identifying two different sequences
as similar. Consider these sequences: $s1$. $\mathtt{ABC}$, $s2$. 
$\mathtt{ABCABCABC}$, and $s3$. $\mathtt{ABC\textcolor{red}{DEFGHI}}$. Most traditional subsequence
matching methods will render $s1$ similar to both $s2$ and $s3$. However, we
call similarity of $s1$ and $s3$ a false positive because $s3$'s overall pattern
is significantly different from $s1$.

In this paper, a new sequence feature embedding
function, Sequence Graph Transform (SGT),
that extracts the short- and long-term sequence features without any increase
in the computation has been developed. This unique property of SGT removes the computation
limitations; it enables us to tune the amount of short- to long-term patterns
that will be optimal for a given sequence problem. SGT also addresses the issue
of false positives upon comparing sequences of different lengths. 

SGT embedding is a nonlinear transform of the inter-symbol distances in a sequence. Its name is attributed to the graphical interpretation of the embedding which shows the ``association'' between sequence symbols.

SGT is in a finite-dimensional feature space that can be used as a vector
in most mainstream data mining methods, such as kmeans, kNN, SVM, and Deep Learning
architectures. Moreover, it can be used as a graph for applying graph mining methods and interpretable visualizations.

We show that these properties have led to a significantly higher accuracy in sequence modeling with lower computation. We theoretically prove the SGT properties,
and experimentally and practically validate its efficacy. We also show that SGT
features can be used as an embedding layer in a Feed-forward Neural Network (FNN). In our
real-world data sets, this outperformed the current state-of-the-art long- and short-term neural network (LSTM) 
classifiers in both runtime and accuracy. 

\subsection{Related Work\label{subsec:related-work}}

Sequence mining is an extensively studied problem. Several works
have been done specifically to estimate sequence similarity and feature representations
for sequence classification, clustering, etc. We categorize the literature
as follows.

\paragraph{\textbf{Alignment.}}

Sequence alignment has two broad types: global alignment \citep{needleman1970general} and local alignment \citep{stoye1997dca}. Global alignment finds the sequence similarity between sequences over their entire length. While they work better in pairwise sequence comparison, it becomes prohibitively time intensive on large sequence data sets.  For them, multiple sequence alignment (MSA) techniques were developed.

Several MSA techniques accomplished global alignment  \citep{notredame2000t, thompson1994improved}. But they were ineffective when sequences have common patterns (homologous) only over local regions. In such cases, local alignment needs to be performed \citep{bailey1994fitting, lawrence1993detecting, morgenstern1999dialign}.

In most of these methods, the computation complexity remains an issue. Dynamic Programming (DP) has been used in the MSA techniques. Here DP suffers from high-dimensional problems in MSA because the number of sequences is equal to the number of dimensions. It is stated in \citet{wang1994complexity} if two or more optimal paths are available and need to trace backward, the complexity of the backtracking grows exponentially. 

MSA is an NP-complete problem. To solve them, two types of approaches are prevalent: exact and progressive alignment. Exact algorithms usually deliver high-quality alignment that is very close to the optimal but applying them on most real problems is unrealistic due to excess complexity \citep{lipman1989tool,stoye1997dca}. Progressive alignment is used in CLUSTAL \citep{thompson1994clustal}, BLAST \citep{altschul1997gapped}, FASTA \citep{pearson19905}, UCLUST \citep{edgar2010search}, CD-HIT \citep{fu2012cd}, and MUSCLE
\citep{edgar2004muscle}. These alignment algorithms are greedy in nature. This does not allow modification of string gaps and, hence, the alignment similarity cannot be adjusted at a later stage. Also, a greedy algorithm can be trapped in local minima. Another major drawback is that most progressive alignments are sensitive to the initialization (the initial alignment). Moreover, most alignment algorithms are heuristics and face these challenges. They, therefore, suffer from accuracy and computation issues due to which sequence alignment is still under research.

\paragraph{\textbf{Kernels.}}

Sequence mining using string kernels has been considerably worked on. In the current literature, kernel function has proven to be an effective method \citep{leslie2004mismatch, xing2010brief}. 

Over the last few decades, several string kernel methods have been proposed, e.g.,  \citet{cristianini2000introduction, kuang2005profile, leslie2001spectrum, eskin2003mismatch, leslie2004mismatch, smola2003fast}. Among them, the k-spectrum kernel \citep{leslie2001spectrum}, (k,m) mismatch kernel, and their variants \citep{eskin2003mismatch, leslie2004mismatch} gained popularity in the early 2000s.

These kernels decompose the original strings into sub-structures, i.e., k-mers (small strings). They then find the count of the k-mers with up to m mismatches in the original sequence to define a feature map. However, only the patterns of short subsequences are captured in these methods. They fail to capture long-term patterns. To address this, if larger k and m are taken, the feature map and the computation grows exponentially. This makes them applicable to only small k, m, and eventually to small data sets \citep{wu2019efficient}.

A thread of recent research has made valid attempts to improve the computation of the kernel matrix, e.g., \citet{farhan2017efficient, kuksa2009scalable}. But these methods only address the scalability issue in terms of the length of strings and the size of the symbols set. The kernel matrix construction still has a quadratic complexity with respect to the number of strings. Moreover, these methods inherit the issues of ``local'' kernels, i.e., long-term dependencies are ignored. 

More recently, a string kernel with random features was introduced \citep{wu2019efficient}. This family of string kernels is defined through a series of different random feature maps. They discover global long-term patterns and maintain a computation cost linear with respect to the string length and the number of string samples. This kernel produces random string embeddings (RSE) by utilizing random feature approximations from randomly generated strings. These random strings have a short length to reduce the computation complexity from quadratic to linear. But the approximation for computational gain has a counter effect on the efficacy of the kernel embeddings.

Another class of kernel methods computes pairwise sequence similarity using some global or local alignment measure, e.g., \citet{needleman1970general, smith1981comparison}. These string alignment kernels are defined using a learning methodology R-convolution \citep{haussler1999convolution}---a framework for computing the kernels between discrete objects. It works by recursively decomposing structured objects into sub-structures and computes their global and local alignments to derive a feature map. While these methods cover both short-term (local) and long-term (global) dependencies, they have high computation costs: quadratic in both the number and the length of sequences.


\paragraph{\textbf{Time-series Classification.}}

Sequences are a special type of time series. A typical time series is a sequence of observations of a continuous variable and a sequence is the same for a discrete or categorical variable. It has been also stated in \citep{gamboa2017deep} that any classification problem in which the data is registered taking into account some notion of order can be cast as time series classification (TSC) problem.

TSC has been deeply studied. With the increase in temporal data, several TSC algorithms have been proposed in the past decade, e.g., \citet{bagnall2015time}. One of the most popular and traditional TSC approaches is the use of the nearest neighbor (NN) classifier in conjunction with a distance function \citep{lines2015time}. \citet{bagnall2015time} showed that dynamic time warping (DTW) distance with an NN classifier worked effectively.

Prior research in \citet{lines2015time} also showed that DTW distance measure worked better than other distance measures. Some recent contributions include ensembling methods, e.g., \citet{bagnall2015time, hills2014classification, bostrom2017binary, lines2016hive, schafer2015boss, kate2016using, deng2013time, baydogan2013bag}. Regardless of the approach, they relied on an effective distance measure. And most of the distance measures faced issues related to effectively capturing both short and long term dependencies with tractable computation.

\paragraph{\textbf{Deep Learning.}}

Deep learning-based methods have touched on a variety of problems including sequence problems. Natural language processing and speech recognition problems solved with deep learning architectures have a similar construct as a sequence problem. Due to this, the methods developed for the former have been adopted in sequence problems, e.g., in \citet{lines2016hive, lines2018time, bagnall2017great, neamtu2018generalized}.

More specifically, LSTMs in RNNs are extensively
used for sequence mining problems  due to its ability to learn
long- and short- term sequence patterns \citep{graves2013generating}. LSTMs are
commonly used for building supervised sequence models and
sequence-to-sequence predictions \citep{sutskever2014sequence}.
However, LSTMs and other RNNs cannot differentiate between length sensitive
and insensitive sequence problems (discussed in \S\ref{subsec:Solution}). The LSTM
layer is also not interpretable for a visualization. Additionally, an LSTM
is computationally intensive compared to an FNN model used with an SGT embedding (shown in \S\ref{subsec:classification}).

\paragraph{\textbf{Pattern discovery.}}

\textit{N}-gram (also known as \textit{k}-mers) methods
and their variants \citep{comin2012alignment,didier2012variable}
are popular approaches for pattern discovery. However,
their feature space and computation increase exponentially for long-term
dependencies. 
Another class of methods does \textit{frequent} subsequence discovery using
\textit{apriori}-like breadth-first search methods or pattern-growth
depth-first search methods, e.g. GSP \citep{srikant1996mining}, PSP \citep{masseglia1998psp}, and SPADE \citep{zaki2001spade}.
These methods, however, had a critical nontrivial computation that was addressed
by PrefixSpan \citep{han2001prefixspan,chiu2004efficient}, and SPAM \citep{ayres2002sequential}. While some of these methods are more suitable for sequences of item sets, most of their feature representations can lead to poor accuracy.

\citet{wang2005introduction} extracted features from protein sequences using a 2-gram encoding method and 6-letter exchange group methods to find the global similarity. They used this with a neural network model. Some user-defined variables like \textit{len}, \textit{mut}, and \textit{occur} were also used to find the local similarities. \citet{wu2006universal} enlarged 2-gram encoding to an n-gram to improve the accuracy. \citet{zainuddin2008radial} developed a radial-based approach to reduce the computational overhead of n-gram encoding method. \citet{zaki2004features} used a hidden Markov model to extract features that were applied to the classifier that can train the data in high-dimensional space. They used their features in building SVM classifiers.

\paragraph{\textbf{Hash maps.}}

Hash maps address the high-dimensional
input spaces for fixed or variable length \textit{n}-gram spaces
by performing dimensionality reduction. They are typically developed in Bioinformatics \citep{buhler2001efficient,buhler2002finding,indyk1998approximate,wesselink2002determining}. \citet{shi2009hash} 
used hashing to compare all subgraph pairs on biological graphs.
However, feature hashing can result in significant loss of information,
especially when hash collisions occur between highly frequent features
with significantly different class distributions. On the lines of
hashing methods, Genome fingerprinting methods 
have  been developed \citep{glusman2017ultrafast}. They are fast and 
accurate. However, they are particularly built for and suitable to
genome data due to the small symbol set size and also known
domain knowledge.

\paragraph{\textbf{Generative.}}
The parametric generative methods typically make Markovian distribution assumptions,
more specifically a first-order Markov property, e.g., \citet{cadez2003model,ranjan2015impact}. However, such a distributional
assumption is not always valid. A general \emph{n-}order Markov model
was also proposed but not popular in practice due to high computation.
 Hidden Markov model-based approaches are popular
in both bioinformatics and general sequence problems \citep{helske2017mixture,remmert2012hhblits}. It assumes a hidden layer
of latent states which results in the observed sequence. These hidden
states have a first-order Markov transition assumption that due to
the multi-layer setting, the first-order assumption is not transmitted
to the observed sequence. However, tuning HMM (finding optimal hidden
states) is difficult and it is computationally intensive.

\paragraph{\textbf{Graph based.}}
Temporal graphs is a category of graph representations similar to SGT defined
in this paper. Temporal
graphs were used for Phenotyping in \citet{liu2015temporal}
and Temporal Skeletonization in \citet{liu2016temporal}. However, the definition
of the developed SGT is fundamentally different from these Temporal graphs. Moreover, SGT's ability to capture the short- and long-term features are theoretically substantiated.
Another class of graph methods hypothesizes that sequences are generated
from some evolutionary process where a sequence is produced by reproducing
complex strings from simpler substrings, as in \citet{siyari2016lexis} and
references therein. However, the estimation algorithms for these methods
are heuristics, sometimes greedy, and have identifiability
issues. Moreover, the evolutionary assumption may not be always true.

\subsection{Research Specification\label{subsec:Solution}}

\subsubsection{Problem}

The related methods discussed above fail to address at least one
of the following challenges: a) capturing long-term dependencies, b) false positives upon comparing sequences of different lengths, and c) domain-specific and/or computation
complexity with respect to
sequence length, sample size, and the size of symbols set, where sequence length is the total number of symbols in the sequence, sample size is the number of sequences in the data set, and the symbols set is the set of unique symbols that make the sequences of the data set.

We propose a new sequence feature extraction function, Sequence
Graph Transform (SGT), that addresses the above challenges
and is shown to outperform existing state-of-the-art methods in sequence
data mining. SGT works by quantifying the pattern in a sequence by
scanning the positions of all symbols relative to each other. We
call it a \emph{graph }transform because of its inherent property
of interpretation as a graph, where the \textit{symbols} form the nodes and
a directed connection between two nodes shows their ``association.''
These ``associations'' between all symbols represent the signature
features of a sequence. 

Sequence analysis problems can be broadly divided into
a) \emph{length-sensitive}: the inherent patterns, as well as the sequence
lengths, should match to render two sequences as similar, e.g., in
protein sequence clustering, and b) \emph{length-insensitive}: the
inherent patterns should be similar, irrespective of the lengths,
e.g., weblog comparisons. In contrast with the existing literature,
SGT provides a solution for both scenarios. The advantage of this
property becomes more pronounced when we have to perform both types
of analysis on the same data and implementing different methods for
each becomes cumbersome. 

\subsubsection{Contribution}

In this paper, our major contribution is the development of a new
sequence feature embedding function: Sequence Graph Transform. SGT embedding exhibits the following properties,

\begin{enumerate}
    \item \textbf{Short- and long-term.} Captures both short and long term dependencies, i.e., both local and global patterns. The amount of long term dependency to incorporate can be controlled with a tuning parameter. Importantly, unlike the existing methods, enlarging the long-term dependency does not affect on SGT computation. By removing the computation limitation, SGT embedding can be effectively tuned based on a problem requirement.
    
	\item \textbf{Computationally tractable.} SGT computation complexity is linear with respect to the number of sequences. Moreover, there are two algorithms proposed for SGT estimation. One is selected based on which is higher between the sequence length and alphabet set.
    
    \item \textbf{Compatibility.} Compatible with mainstream supervised and unsupervised learning methods. SGT is an embedding function that converts an unstructured sequence into a finite-dimensional vector. Mainstream learning methods, such as SVM classifier or k-means clustering, take such vectors as inputs. SGT embedding used in conjunction with mainstream learning methods yields significantly superior results for sequence problems.
    
    \item \textbf{Interpretability}. Embedded features are interpretable. Each feature in an SGT embedding corresponds to a directional dependency between a symbol pair. For example, SGT embedding of a sequence $\mathtt{BAABCCA}$ will have a feature corresponding to each 2-permutation of symbols: $\mathtt{(A,A);\, (A,B);\,}$ $\mathtt{(A,C);\, (B,A);\, (B,B);\, (B,C);\, (C,A);\, (C,B);\, (C,C)}$. The symbol order in a feature $\mathtt(i,j)$ indicates the forward dependency from $\mathtt{i}$ to $\mathtt{j}$: a high value of feature indicates a high forward dependency, i.e., $\mathtt{i}$ is followed by a significant amount of $\mathtt{j}$'s in the sequence.
    
\end{enumerate}{}

It is important to note that SGT is an embedding methodology. It is used in conjunction with mainstream supervised and unsupervised learning methods. The source code, data sets, and illustrative examples are provided at \url{https://github.com/cran2367/sgt} (see Appendix~\ref{sec:code-repository-and-data-sets}).

\subsubsection{Limitations}

SGT works in most sequence problems but has the following limitations.

\begin{enumerate}
    \item The effectiveness of SGT embedding becomes diminished if the sequence symbol set is small. For example, in binary or DNA sequences where $symbols \in \{0,1\}$ and $\{\mathtt{A,C,G,T}\}$, respectively. This is because the size of the embedding is proportional to the size of the symbol set. In the above examples, the SGT embedding will be of size 4 and 16, respectively. But if the sequence length is high, the embedding cannot hold sufficient information that characterizes the sequence.
    
    \item The proposed SGT algorithm applies to sequences of single element items called as \textit{symbols}. The reason is that SGT works by extracting dependencies between items in a sequence. For this, SGT assumes an item to be unique. However, an item is a bucket of elements---items can have common elements, e.g., the items in $\{\langle B \rangle
\langle AAB \rangle \langle CC \rangle \langle A \rangle \}$ share elements. The proposed SGT does not draw information from the presence of common elements in items. An extension to multi-element item sequences is non-trivial and should be pursued in future research.
\end{enumerate}{}


In the following,
we develop SGT and provide its theoretical support. We
perform an extensive experimental evaluation and show that SGT bridges the gap between
sequence mining and mainstream data mining through direct application
of fundamental methods, viz. PCA, k-means, SVM, and graph visualization
via SGT on sequence data analysis.

\section{Sequence Graph Transform (SGT)}
\label{sec:Sequence-Graph-Transform}

\subsection{Overview and Intuition}
\label{subsec:SGT-Overview}

\begin{figure}
\begin{centering}
\includegraphics[scale=0.3]{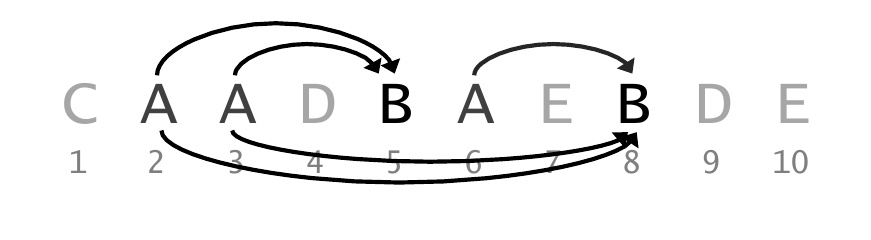}
\par\end{centering}
\caption{Illustration of the ``effect'' of elements on each other.\label{fig:Illustration-of-effect}}
\end{figure}

By definition, a sequence can be either feed-forward or bidirectional.
In a feed-forward sequence, events (symbol instances) occur in succession;
e.g., in a clickstream sequence, the click events occur one after another
in a forward direction. On the other hand, in a bidirectional
sequence, the directional or chronological order of symbol instances
is not present or not important. In this paper, we present SGT for 
feed-forward sequences; SGT for bidirectional sequences is given in \S~\ref{subsec:sgt-for-undirected-sequences}.

For either of these sequence types, the developed SGT
works on a fundamental premise\textemdash{}the relative positions
of symbols in a sequence characterize the sequence\textemdash{}to extract the pattern features of the sequence. This premise holds for most sequence mining problems because the similarity in sequences
is often measured based on the similarities in their pattern from the symbol positions. 

Fig.~\ref{fig:Illustration-of-effect} shows
an illustrative example of a feed-forward sequence. In this example, the presence of symbol $\mathtt{B}$ at positions 5 and 8 should
be seen in context with or as a result of all other predecessors.
To extract the sequence features, we take the relative positions of one
symbol pair at a time. For example, the relative positions for pair
($\mathtt{A}$,$\mathtt{B}$) are \{(2,3),5\} and \{(2,3,6),8\}, where
the values in the position set for $\mathtt{A}$ are the ones preceding
$\mathtt{B}$. In the SGT procedure defined and developed in the following
sections (\S\ref{subsec:SGT-definition}-\ref{subsec:SGT-properties}),
the sequence features are shown to be extracted from these positions
information.

\begin{figure}
\centering     
\subfloat[Feature embedding in a vector with a graph interpretation.]
		{\label{fig:SGT-overview-a-one-seq}\includegraphics[width=70mm]{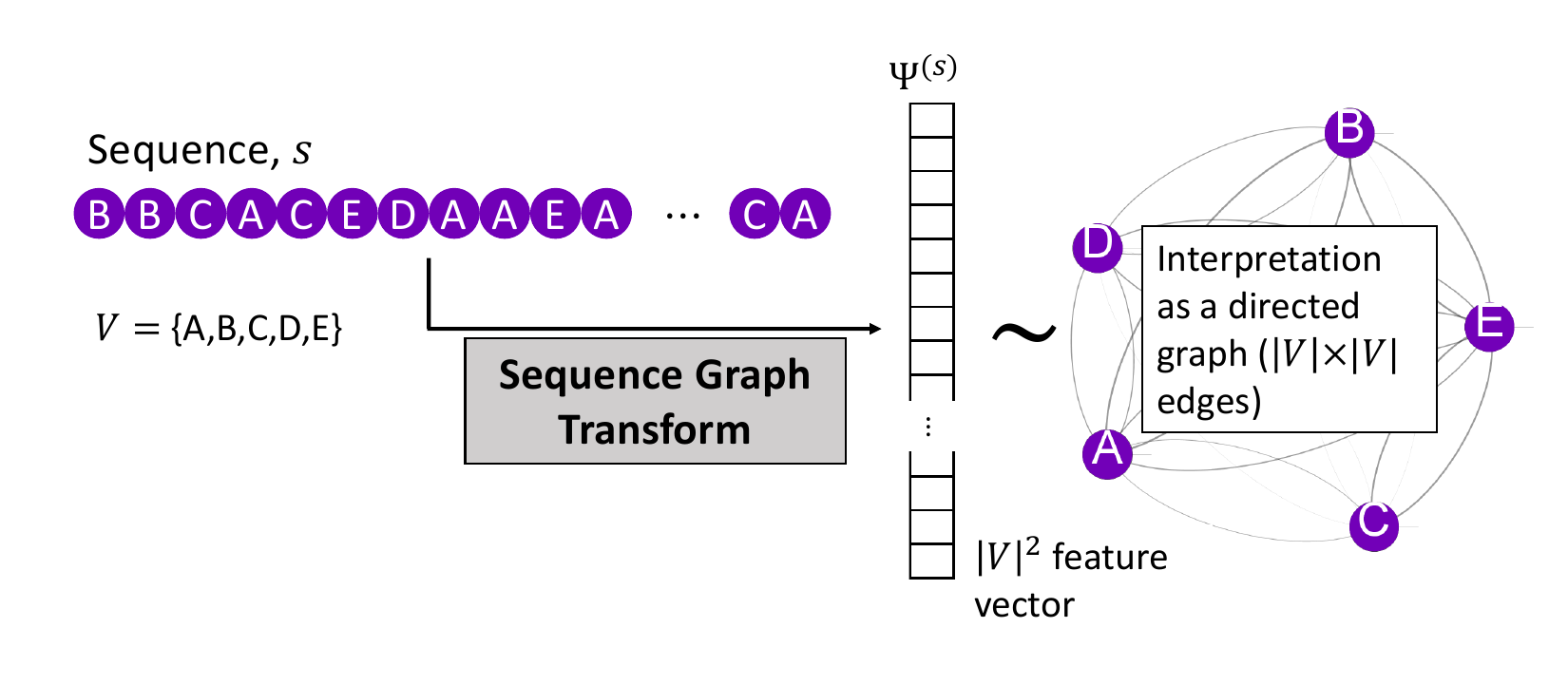}}
\newline
\subfloat[Use of sequences' SGT embedding for data mining.]
		{\label{fig:SGT-overview-b-seq-data}\includegraphics[width=70mm]{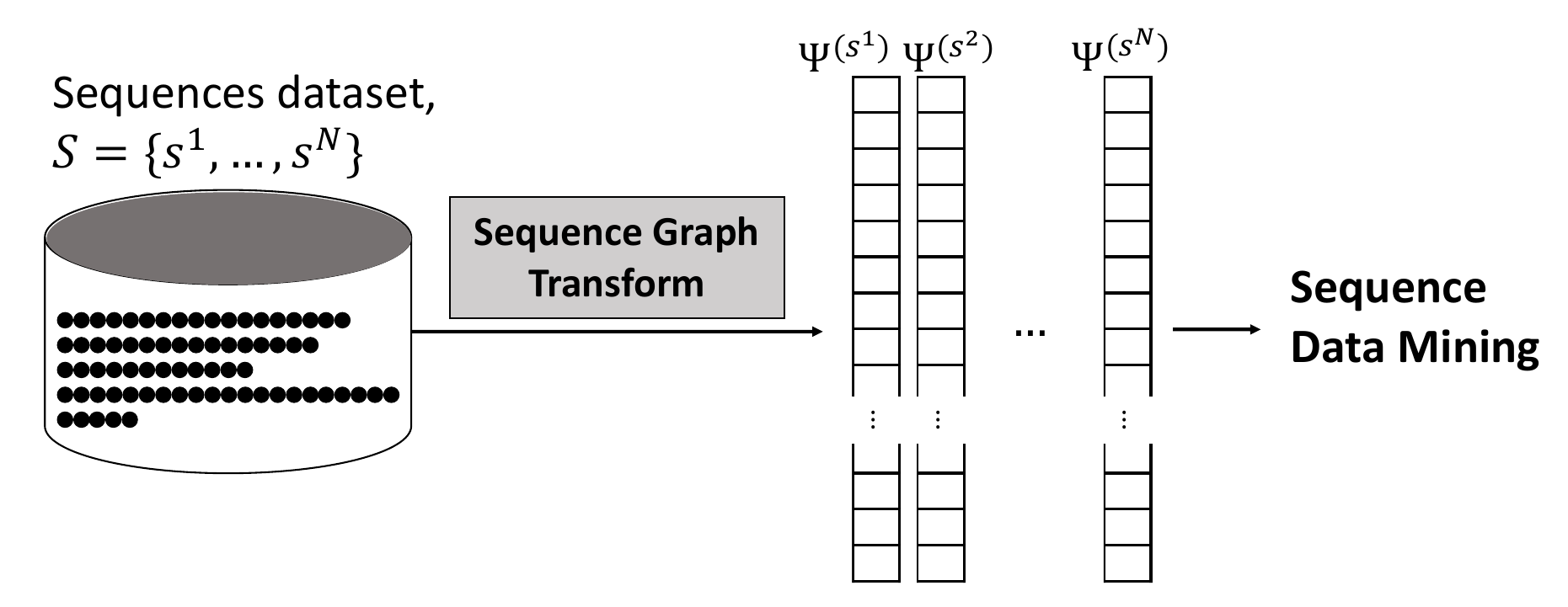}}

\caption{SGT overview.}
\end{figure}

These extracted features are an ``association'' between $\mathtt{A}$
and $\mathtt{B}$, which can be interpreted as a connection feature
representing ``$\mathtt{A}$ leading to $\mathtt{B}$.'' We should
note that ``$\mathtt{A}$ leading to $\mathtt{B}$'' will be different
from ``$\mathtt{B}$ leading to $\mathtt{A}$.'' The associations between all symbols in the symbol
set denoted as $\mathcal{V}$ can be extracted similarly to obtain
sequence features in a $|\mathcal{V}|^{2}$-dimensional space. 

This is similar to the Markov probabilistic models, in which the transition probability
of going from $\mathtt{A}$ to $\mathtt{B}$ is estimated. However,
SGT is different because the connection feature 1) is not a probability,
and 2) takes into account all orders of the relationship without any increase
in computation. 

Besides, the SGT also make it easy to visualize the sequence as a
directed graph, with sequence symbols in $\mathcal{V}$ as graph
nodes and the edge weights equal to the directional association
between nodes. Hence, we call it a sequence \emph{graph} transform.
Moreover, we show in \S~\ref{subsec:sgt-for-symbol-clustering} that under certain conditions, the SGT also allows
node (symbol) clustering.

A high-level overview of our approach is given in Fig.~\ref{fig:SGT-overview-a-one-seq}-\ref{fig:SGT-overview-b-seq-data}.
In Fig.~\ref{fig:SGT-overview-a-one-seq}, we show that applying SGT
on a sequence, $s$, yields a finite-dimensional  SGT feature vector $\Psi^{(s)}$
for the sequence, also interpreted and visualized as a directed graph.
For a general sequence data analysis, SGT can be applied to each sequence
in the sample (Fig.~\ref{fig:SGT-overview-b-seq-data}). 
The resulting feature vectors can be used with mainstream
data mining methods.

\subsection{Notations\label{subsec:Notations}}

Suppose we have a data set of sequences denoted by $\mathcal{S}$.
Any sequence in the data set, denoted by $s$($\in\mathcal{S}$), is
made of symbols in set $\mathcal{V}$. A sequence can have instances
of one or many symbols from $\mathcal{V}$. For example, sequences
from a data set, $\mathcal{S}$, made of symbols in $\mathcal{V}=\{\mathtt{A,B,C,D,E}\}$(suppose)
can be $\mathcal{S}=\{\mathtt{AABAAABCC,DEEDE}$, $\mathtt{ABBDECCABB,}\ldots\}$. The length of a sequence, $s$, denoted by, $L^{(s)}$,
is equal to the number of events (in this paper, the term ``event'' is used for a symbol instance) in it. In the sequence, $s_{l}$
will denote the symbol at position $l$, where $l=1,\ldots,L^{(s)}$
and $s_{l}\in\mathcal{V}$. 

We extract a sequence $s$'s features
in the form of ``associations'' between the symbols, represented
as $\psi_{uv}^{(s)}$, where $u,v\in\mathcal{V}$, are the corresponding
symbols, and $\psi$ is a function of a helper function $\phi$. $\phi_{\kappa}(d)$ is a
function that takes a ``distance,'' $d$, as input, and $\kappa$
as a tuning hyper-parameter. 

\subsection{SGT Definition\label{subsec:SGT-definition}}

\begin{figure}
	\centering
    \subfloat[]{\label{fig:phi-individual-effect}
    \includegraphics[scale=0.3]{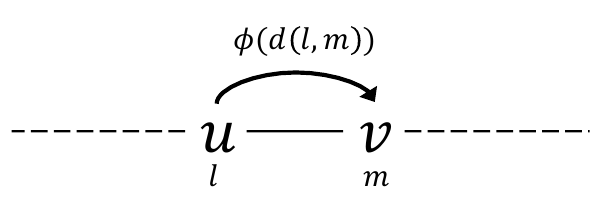}}
    \subfloat[]{\label{fig:phi-general-effect}
    \includegraphics[scale=0.26]{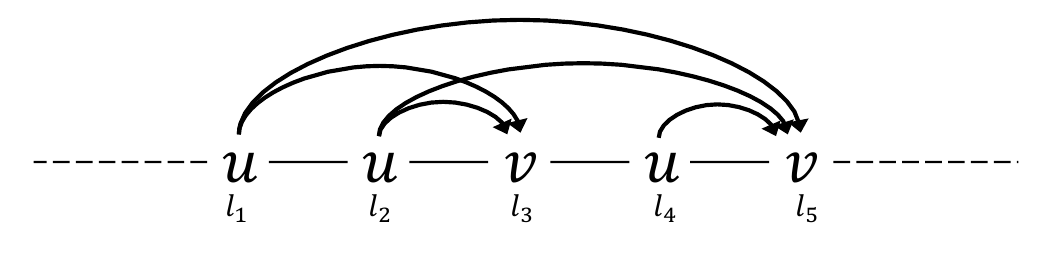}}
     
\caption{Illustration of the effect of symbols' relative positions.}
\label{fig:phi-illustration}
\end{figure}

As also explained in \S\ref{subsec:SGT-Overview}, SGT extracts the features from the relative positions
of events. A quantification for an ``effect''
from the relative positions of two events in a sequence is given by $\phi(d(l,m))$,
where $l,m$ are the positions of the events, and $d(l,m)$ is a distance
 measure. This quantification is an effect of the
preceding event on the later event. For example, see Fig.~\ref{fig:phi-individual-effect},
where $u$ and $v$ are at positions $l$ and $m$, and the directed
arc denotes the effect of $u$ on $v$.

For developing SGT, we require the following conditions on $\phi$: a)
strictly greater than 0: $\phi_{\kappa}(d)>0;\:\forall\,\kappa>0,\,d>0$;
b) strictly decreasing with $d$: $\frac{\partial}{\partial d}\phi_{\kappa}(d)<0$;
and c) strictly decreasing with $\kappa$: $\frac{\partial}{\partial\kappa}\phi_{\kappa}(d)<0$.

The first condition is to keep the extracted SGT feature, $\psi=f(\phi)$,
easy to analyze, and interpret. The second condition strengthens the
effect of closer neighbors. The last condition helps in tuning the
procedure, allowing us to change the effect of neighbors. 

There are several functions that satisfy the above conditions: e.g.,
Gaussian, Inverse and Exponential. We take $\phi$
as an exponential function because it will yield interpretable results
for the SGT properties (\S\ref{sec:sgt-properties-short-long-term-features}) and $d(l,m)=|m-l|$.


\begin{equation}
\phi_{\kappa}(d(l,m))=e^{-\kappa|m-l|},\:\forall\,\kappa>0,\,d>0\label{eq:phi-expression}
\end{equation}

In a general sequence, we will have several instances of a symbol
pair. For example, see Fig.~\ref{fig:phi-general-effect}, where
there are five $(u,v)$ pairs, and an arc for each pair shows an effect
of $u$ on $v$. Therefore, the first step is to find the number of
instances of each symbol pair. The instances of symbol pairs are
stored in a $|\mathcal{V}|\times|\mathcal{V}|$ asymmetric matrix,
$\Lambda$. Here, $\Lambda_{uv}$ will have all instances of symbol
pairs $(u,v)$, such that in each pair instance, $v$'s position is
after $u$.


\begin{eqnarray}
\Lambda_{uv}(s) & = & \{(l,m):\,s_{l}=u,s_{m}=v,\nonumber \\
 &  & l<m,l,m\in1,\ldots,L^{(s)}\}
 \label{eq:psi-lambda-main}
\end{eqnarray}


After computing $\phi$ from each $(u,v)$ pair instance for the sequence,
we define the ``association'' feature $\psi_{uv}$ as a normalized
aggregation of all instances, as shown below in Eq.~\ref{eq:psi-main-len-sensi}-\ref{eq:psi-main-len-insensi}. 
Here, $|\Lambda_{uv}|$ is the size of the set
$\Lambda_{uv}$, which is equal to the number of $(u,v)$ pair instances.
Eq.~\ref{eq:psi-main-len-sensi} gives the feature expression for
a \emph{length-sensitive }sequence analysis problem because it also
contains the sequence length information within it (proved with a
closed-form expression under certain conditions in
\S\ref{sec:sgt-properties-short-long-term-features}). In Eq.~\ref{eq:psi-main-len-insensi},
the length effect is removed by normalizing $|\Lambda_{uv}|$ with
the sequence length $L^{(s)}$ for \emph{length-insensitive }problems 
(shown in \S\ref{sec:sgt-properties-short-long-term-features}). 

\begin{subnumcases}{\psi_{uv}(s)=}
   \cfrac{\sum_{\forall(l,m)\in\Lambda_{uv}(s)}e^{-\kappa |m-l|}}{|\Lambda_{uv}(s)|} & $\textnormal{;\,length\,sensitive}$ \label{eq:psi-main-len-sensi}
   \\
   \cfrac{\sum_{\forall(l,m)\in\Lambda_{uv}(s)}e^{-\kappa |m-l|}}{|\Lambda_{uv}(s)|/L^{(s)}} & $\textnormal{;\,length\,insensitive}$ \label{eq:psi-main-len-insensi}
\end{subnumcases}

and $\Psi(s)=[\psi_{uv}(s)],\:u,v\in\mathcal{V}$ is the SGT feature
representation of sequence $s$.

For illustration, the SGT feature for symbol pair $\mathtt{(A,B)}$
in sequence in Fig.~\ref{fig:Illustration-of-effect} can be computed
as (for $\kappa=1$ in length-sensitive SGT): $\Lambda_{\mathtt{AB}}$$=$ $\{(2,5);(3,5);$ $(2,8);$ $(3,8);$ $(6,8)\}$
and $\psi_{\mathtt{AB}}$$=$$\frac{\sum_{\forall(l,m)\in\Lambda_{\mathtt{AB}}}e^{-|m-l|}}{|\Lambda_{\mathtt{AB}}|}= \\ \frac{e^{-|5-2|}+e^{-|5-3|}+e^{-|8-2|}+e^{-|8-3|}+e^{-|8-6|}}{5}=0.066$.

The features, $\Psi^{(s)}$,
can be either interpreted as a directed ``graph,'' with edge weights, $\mathcal{\psi}$,
and nodes in $\mathcal{V}$ or vectorized to a $|\mathcal{V}|^{2}$-vector
denoting the sequence $s$ in the feature space. 

\subsection{SGT Properties}
\label{subsec:SGT-properties}

\subsubsection{Short- and long-term Features}
\label{sec:sgt-properties-short-long-term-features}

In this section, we show SGT's property of capturing both short- and
long-term sequence pattern features. This is shown by a closed-form
expression for the expectation and variance of the SGT feature, $\psi_{uv}$, under some assumptions. 

\begin{figure}
\begin{centering}
\includegraphics[scale=0.31]{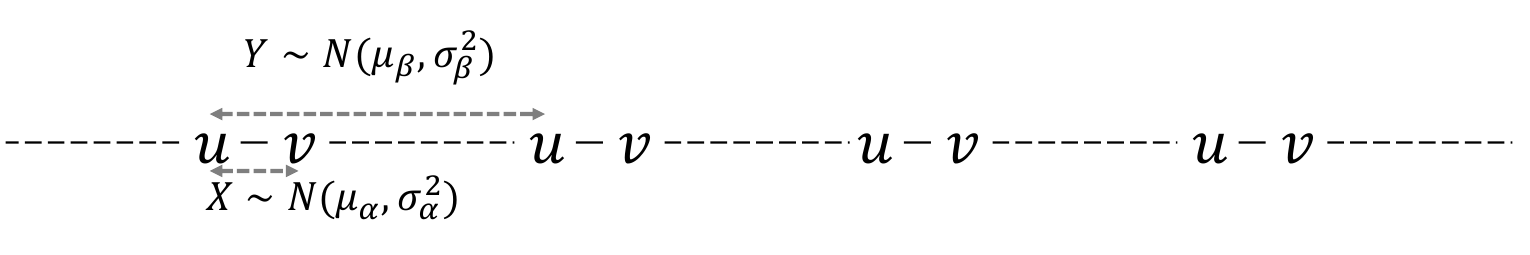}
\par\end{centering}
\caption{Representation of short- and long-term dependencies.\label{fig:Representation-of-short-long}}
\end{figure}

Assume a sequence of length $L$ with an inherent pattern: $u$,$v$
occur closely together within a stochastic gap as $X\sim N(\mu_{\alpha},\sigma_{\alpha}^{2})$,
and the intermittent stochastic gap between the pairs as $Y\sim N(\mu_{\beta},\sigma_{\beta}^{2})$,
such that, $\mu_{\alpha}<\mu_{\beta}$ (See Fig.~\ref{fig:Representation-of-short-long}). $X$ and $Y$
characterize the short- and long-term patterns, respectively.
Note that this assumption is only for showing
an interpretable expression and is not required in practice.


\begin{theorem}
\label{theorem:closed-form}
	The expectation and variance of SGT feature, $\psi_{uv}$, has a closed-form expression under the above assumption, which
shows that it captures both short- and long-term patterns present
in a sequence in both length- sensitive and insensitive SGT variants. 



\begin{align}
E[\psi_{uv}] & = & \begin{cases}
\cfrac{2}{pL+1}\gamma & ;\text{length sensitive}\\
\cfrac{2L}{pL+1}\gamma & ;\text{length insensitive}
\end{cases}\label{eq:E-phi-AB-part-2-2}
\end{align}

\begin{align}
\textnormal{var}(\psi_{uv}) & = & \begin{cases}
\left(\cfrac{1}{pL(pL+1)/2}\right)^{2}\pi & ;\text{length sensitive}\\
\left(\cfrac{1}{p(pL+1)/2}\right)^{2}\pi & ;\text{length insensitive}
\end{cases}
\label{eq:Var-phi-AB-part-2-2}
\end{align}

\vspace{-0.2in}

where, 

\vspace{-0.25in}

\begin{align}
\gamma & = & \frac{e^{-\tilde{\mu}_{\alpha}}}{\left|\left(1-e^{-\tilde{\mu}_{\beta}}\right)\left[1-\frac{1-e^{-pL\tilde{\mu}_{\beta}}}{pL(e^{\tilde{\mu}_{\beta}}-1)}\right]\right|}
\label{eq:weight-constant-lambda-1}
\end{align}
\begin{align}
\pi & = & \cfrac{e^{-2\tilde{\mu}{}_{\alpha}}}{1-e^{-2\tilde{\mu}{}_{\beta}}}\left(pL-e^{-2\tilde{\mu}_{\beta}}\left(\cfrac{1-e^{-2pL\tilde{\mu}{}_{\beta}}}{1-e^{-2\tilde{\mu}{}_{\beta}}}\right)\right)
\label{eq:weight-constant-pi-1}
\end{align}

and, $\tilde{\mu}_{\alpha}=\kappa\mu_{\alpha}-\frac{\kappa^{2}}{2}\sigma_{\alpha}^{2};\tilde{\mu}_{\beta}=\kappa\mu_{\beta}-\frac{\kappa^{2}}{2}\sigma_{\beta}^{2}$, $p\rightarrow \text{constant}$.

\end{theorem}

\textit{Proof.} Given in Appendix~\ref{appendix:theorem-mean-variance}.

As we can see in Eq.~\ref{eq:E-phi-AB-part-2-2},
the expected value of the SGT feature is proportional to the term
$\gamma$. The numerator of $\gamma$
contains information about the short-term pattern, and its denominator
has long-term pattern information.

In Eq.~\ref{eq:weight-constant-lambda-1}, we can observe that if either of $\mu_{\alpha}$
(the closeness of $u$ and $v$ in the short-term) and/or $\mu_{\beta}$
(the closeness of $u$ and $v$ in the long-term) decreases, $\gamma$
will increase, and vice versa. This emphasizes two properties: a)
the SGT feature, $\psi_{uv}$, is affected by changes in both short- and long-term patterns, and b) $\psi_{uv}$ increases when $u,v$ becomes closer in the short- or long- range in the sequence, providing an analytical connection between the observed pattern and the extracted feature. Besides, it also proves the graph interpretation of SGT: $\psi_{uv}$ that denotes the edge weight for nodes $u$ and $v$ (in the SGT-graph) increases if closeness
between $u,v$ increases in the sequence, meaning that the nodes become closer in the graph space (and vice
versa). Importantly, $\lim_{L\rightarrow\infty}\text{var}(\psi_{uv})\rightarrow0$
ensures feature stability.

In the length-insensitive SGT feature expectation in Eq.~\ref{eq:E-phi-AB-part-2-2},
it is straightforward to show that it becomes independent of the sequence
length as the length increases. As sequence length, $L$,
increases, the $(u,v)$ SGT feature approaches a constant, given as $\lim_{L\rightarrow\infty}E[\psi_{uv}] \rightarrow  \cfrac{2}{p}\left|\cfrac{e^{-\tilde{\mu}_{\alpha}}}{1-e^{-\tilde{\mu}_{\beta}}}\right|$.

Besides, for this  $\lim_{L\rightarrow\infty}\text{var}(\psi_{uv})\underset{1/L}{\rightarrow}0$.
Thus, the expected value of the SGT feature becomes independent of
the sequence length at a rate of inverse to the length. In our experiments,
we observe that the SGT feature approaches a length-invariant
constant when $L>30$.

\vspace{-0.1in}

\begin{eqnarray}
\lim_{L\rightarrow\infty}\Pr\left\{ \psi_{uv}=\cfrac{2}{p}\left|\cfrac{e^{-\tilde{\mu}_{\alpha}}}{1-e^{-\tilde{\mu}_{\beta}}}\right|\right\}  & \underset{1/L}{\rightarrow} & 1\label{eq:E-psi-AB-len-insensi-lim-approaches-1}
\end{eqnarray}

\vspace{-0.1in}

Furthermore, the length-sensitive SGT feature expectation in Eq.~\ref{eq:E-phi-AB-part-2-2}
contains the sequence length, $L$. This shows that the SGT feature
has the information of the sequence pattern, as well as the sequence length.
This enables an effective length-sensitive sequence analysis because
sequence comparisons via SGT will require both patterns and sequence
lengths to be similar.

Additionally, for either case, if the pattern variances, $\sigma_{\alpha}^{2}$ and
$\sigma_{\beta}^{2}$, in the above scenario are small, $\kappa$
allows regulating the feature extraction: higher $\kappa$ reduces
the effect from long-term patterns and vice versa.

\subsubsection{Uniqueness of SGT Sequence
Encoding}

The properties discussed above play an important role in SGT's effectiveness.
Due to these properties, unlike the methods discussed in \S\ref{subsec:related-work},
SGT can capture higher orders of relationships without any increase
in computation. Besides, SGT can effectively find sequence features
without the need for any hidden string/state(s) search.

In this section, we show an additional property of SGT useful for
sequence encoding while answering, is SGT feature for a sequence unique? Yes and no. Based on Theorem~\ref{theorem-sgt-stack} 
given below, a stack of SGTs computed for sufficiently different values
of $\kappa$ will be a unique representation of a sequence. This representation
can also be used for sequence encoding.

However, in a typical sequence mining problems, we require sequences
with similar (same) patterns to be close (equal) in its feature space.
This makes data separation in clustering and boundary computation in 
classification easier. Therefore, stacking SGTs is usually not required
as also found in our results.

\begin{theorem}\label{theorem-sgt-stack}
A stack of SGTs for a sequence $s$, $\Psi^{(\kappa)}(s), \kappa = 1,2,\ldots$ uniquely characterizes the sequence. 
\end{theorem}

\begin{proof}
The theorem can be proved if we prove that a sequence $s$ can be reconstructed from SGT components $\mathbf{W}^{(\kappa)}, \kappa = 0, 1,\ldots, K$ with probability 1 as $K \rightarrow \infty$, given its length $L$.

For reconstruction, we have to find the elements present at each position, $x_l, l=1,\ldots, L$.

$\mathbf{W}^{(0)}$ gives the initialization of the number of occurrences of each paired instances of elements $u,v \in \mathcal{V}$.

We solve the following system of equations where the unknowns are, $x_l, l=1,\ldots, L$ using the known $\mathbf{W}^{(\kappa)}, \kappa = 1, 2, \ldots$

\begin{eqnarray}
\sum e^{-\kappa|x_l - x_m|} & = & W^{(\kappa)}_{uv}, u,v\in \mathcal{V} \label{eq:appx-theorem2_2-1}
\end{eqnarray}

Solution of this system of equations will yield multiple solutions for $x_l, l=1,2,\ldots,L$. Suppose the set of solutions after solving the system of Eq.~\ref{eq:appx-theorem2_2-1} for $\kappa=1,\ldots,L$ is $\alpha$.

Since $\mathbf{W}^{(\kappa)}, \kappa = 1, 2, \ldots$ are independent (see proof in Appendix~\ref{appendix:wk-independence-proof}), adding another system of equations for $\kappa = K+1$ will result into a reduced set of solutions $\alpha'$, i.e. $|\alpha| < |\alpha'|$.

Therefore, by induction as $K \rightarrow \infty$, $|\alpha| \rightarrow 1$, i.e. we reach a unique solution which is the reconstructed sequence.

\end{proof}

SGTs can be stacked if the objective is to ensure sequences 
in a data set do not map to the same representation. However, in
most sequence mining problems the objective is to identify
sequences that have similar inherent patterns. To that end, only
one SGT that appropriately captures the long- and short- term
patterns is usually sufficient.

\section{SGT Algorithm}
\label{sec:SGT-Algorithm}

\begin{algorithm}[tb]
   \caption{SGT embedding via sequence parsing.}
   \label{alg:Parsing-Sequence-algo}
\begin{algorithmic}[1]
   \STATE {\bfseries Input:} A sequence, $s\in\mathcal{S}$ , alphabet set, $\mathcal{V}$ , and $\kappa$.
   \vspace{0.1in}
   \STATE \textbf{Initialize} 
   
   \STATE $\mathbf{W}^{(0)},\mathbf{W}^{(\kappa)} \gets \mathbf{0}_{\mathcal{V}\times\mathcal{V}}$, and length, $L \gets 1$.
   \vspace{0.1in}
   \STATE {\bfseries Processing}
   
   \FOR{$i\in\{1,\ldots,(\text{length}(s)-1)\}$}
   
   \FOR{$j\in\{(i+1),\ldots,\text{length}(s)\}$}
   
   \IF{$\text{length-insensitive}\; is \; \text{True}$}
   		\STATE $\mathbf{W}^{(0)}_{s_i,s_j}\gets \mathbf{W}^{(0)}_{s_i,s_j} + 1 / L$
	\ELSE
		\STATE $\mathbf{W}^{(0)}_{s_i,s_j}\gets \mathbf{W}^{(0)}_{s_i,s_j} + 1$
   \ENDIF
   
   \STATE $\mathbf{W}^{(\kappa)}_{s_i,s_j}\gets\mathbf{W}^{(\kappa)}_{s_i,s_j} + \exp(-\kappa|j-i|)$
   \STATE where, $s_i, s_j \in \mathcal{V}$
   \ENDFOR
   \STATE $L \gets L + 1$
   \ENDFOR
   \vspace{0.1in}
   \STATE \textbf{Output} 
   \STATE $\psi_{uv}(s)\gets\left(\frac{W^{(\kappa)}_{u,v}}{W^{(0)}_{u,v}}\right)^{\frac{1}{\kappa}}$;$\Psi(s)=[\psi_{uv}(s)],\:u,v\in\mathcal{V}$

\end{algorithmic}
\end{algorithm}

\begin{algorithm}[tb]
   \caption{SGT embedding via alphabets parsing.}
   \label{alg:Parsing-Sequence-algo-1}
	\begin{algorithmic}[1]
    
   \STATE {\bfseries Input:} A sequence, $s\in\mathcal{S}$ , alphabet set, $\mathcal{V}$ , and $\kappa$.
   \vspace{0.1in}
   \STATE \textit{function} GetAlphabetPositions$(s,\mathcal{V})$
   
		\STATE $\text{positions} \leftarrow \{ \emptyset \} $  
		\FOR{$v\in\mathcal{V}$}
			\STATE $\text{positions}(v)\leftarrow\{i:s_i=v,i=1,\ldots,\text{length}(s)\}$
		\ENDFOR
		\STATE \texttt{return} $\text{positions}$
\vspace{0.1in}
   
   \STATE \textbf{Initialize} 
   \STATE $\mathbf{W}^{(0)},\mathbf{W}^{(\kappa)} \gets \mathbf{0}_{\mathcal{V}\times\mathcal{V}}$, and length, $L \gets 0$ \\$\text{positions}$ $\gets$ GetAlphabetPositions$(s,\mathcal{V})$.

\vspace{0.1in}   
   \STATE {\bfseries Processing}
   
   \FOR{$u\in\mathcal{V}$}
		\STATE $U\leftarrow\text{positions}(u)$
		\FOR{$v\in\mathcal{V}$}
			\STATE $V\leftarrow\text{positions}(v)$
			\STATE $C\leftarrow(U\times V)^{+}=\{(i,j)|i\in U,j\in V,\,\&\,j>i\}$
			
			\IF{$\text{length-insensitive}\; is \; \text{True}$}
				\STATE $\mathbf{W}^{(0)}_{u,v}\leftarrow\text{length}(C) / L$
			\ELSE
				\STATE $\mathbf{W}^{(0)}_{u,v}\leftarrow\text{length}(C)$
			\ENDIF
			
			\STATE $\mathbf{W}^{(\kappa)}_{u,v}\leftarrow\text{sum}(\exp(-\kappa|C_{:,u}-C_{:,v}|))$
		\ENDFOR
		\STATE $L \gets L + \text{length}(U)$
	\ENDFOR

   \vspace{0.1in}
   \STATE \textbf{Output} 
   \STATE $\psi_{uv}(s)\gets\left(\frac{W^{(\kappa)}_{u,v}}{W^{(0)}_{u,v}}\right)^{\frac{1}{\kappa}}$; $\Psi(s)=[\psi_{uv}(s)],\:u,v\in\mathcal{V}$

\end{algorithmic}
\end{algorithm}

We have devised two algorithms for SGT. The first algorithm (see Algorithm~\ref{alg:Parsing-Sequence-algo}) is faster for short sequences when the sequence lengths on average are significantly smaller than the size of the symbol set, i.e., $L<<|\mathcal{V}|$. The second (see Algorithm~\ref{alg:Parsing-Sequence-algo-1}) is faster otherwise. 

The input to the algorithms is a sequence $s$, a symbol set $\mathcal{V}$ that makes up the sequence, and a tuning parameter $\kappa$. The symbol set $\mathcal{V}$ can be larger than the set of symbols present in the sequence $s$, i.e., $\mathcal{V} \supseteq \{s_i\},\, s_i \in s,\, s_i \neq s_j\, \forall i,j$. If the sequence $s$ belongs to a population of sequences $\mathcal{S}$, i.e., $s \in \mathcal{S}$, then the symbol set should comprise of symbols that construct all the sequences in $\mathcal{S}$, i.e., $\mathcal{V} \leftarrow \cup  \{s_i\},\, s_i \in s,\, s_i \neq s_j \, \forall i,j,\, s\in \mathcal{S}$.

The algorithms are initialized with two zero square matrices $\mathbf{W}^{(0)},\mathbf{W}^{(\kappa)}$ of size $|\mathcal{V}|$. Their rows and columns refer to the symbols in $\mathcal{V}$, and a cell will denote a value for the corresponding symbols $(u,v)$. These matrices will be iteratively updated during the SGT computation. They are expected to be sparse if the symbol set is large. Therefore, a sparse representation of $\mathbf{W}^{(0)},\mathbf{W}^{(\kappa)}$ can also be used for computational gains. Besides, a length variable $L$ is initialized to $1$ and $0$ in Algorithm~\ref{alg:Parsing-Sequence-algo} and ~\ref{alg:Parsing-Sequence-algo-1}, respectively. $L$ is also updated during the learning iterations and used if the SGT embedding is for a length-insensitive problem (\verb|if length-insensitive is True|).

$\mathbf{W}^{(\kappa)}$ and $\mathbf{W}^{(0)}$ denote the numerator and denominator in Equation~\ref{eq:psi-main-len-sensi} and \ref{eq:psi-main-len-insensi}, respectively. These terms are computed differently in the two algorithms. In Algorithm~\ref{alg:Parsing-Sequence-algo}, the sequence $s$ is traversed element by element in a double nested $(i,j)$ iterations (lines~3-4). Inside an $(i,j)$ iteration the corresponding symbols $(s_i, s_j)$ are taken from the sequence $s$. For this $(s_i, s_j)$ the cells $\mathbf{W}^{(0)}_{s_i,s_j}$ and $\mathbf{W}^{(\kappa)}_{s_i,s_j}$ are incremented. These are one-step increments for every instance of $(u,v) \forall u,v \in \mathcal{V}$ in $s$.

Differently, instead of traversing the sequence Algorithm~\ref{alg:Parsing-Sequence-algo-1} traverses the symbol set $\mathcal{V}$ in a double nested $(u,v)$ iterations (lines~9-11). This is computationally more efficient for long sequences that have relatively smaller symbols set.

To facilitate this algorithm, a helper function \verb|GetSymbolPositions| is defined. It returns a $\{u:\, \{position\}\}$ dictionary where $u\in \mathcal{V}$ and  $\{position\}$ is the list of indexes at which $u$ is present in $s$.

Inside a $(u,v)$ iteration the positions of the symbols $u$ and $v$ are known. In line~13, the cross product of the positions is taken such that the position of $v$ is after $u$. The constraint is for a feed-forward sequence and can be omitted for a bidirectional sequence.

The cells $\mathbf{W}^{(0)}_{u,v}$ and $\mathbf{W}^{(\kappa)}_{u,v}$ are then updated with the net value of the $(i,j)$ instances in the cross product. Unlike the update steps in Algorithm~\ref{alg:Parsing-Sequence-algo}, the increments here are accumulative.

The sequence length $L$ is computed in the outer iteration loop in both the algorithms. If the embedding is length-insensitive, $\mathbf{W}^{(0)}$ is scaled by the sequence length. Finally, the SGT embedding as the $\kappa$-th root  of the element-wise division of $\mathbf{W}^{(\kappa)},\mathbf{W}^{(0)}$, i.e., $\left(\frac{W^{(\kappa)}_{u,v}}{W^{(0)}_{u,v}}\right)^{\frac{1}{\kappa}}$, is outputted.

The resulting embedding is a $|\mathcal{V}|\times|\mathcal{V}|$ matrix. The matrix can be used as is for feature interpretation, visualization, or learning similar to the way an adjacency matrix is used. The embedding is, otherwise, vectorized to a $|\mathcal{V}|*|\mathcal{V}|$ vector and used as input to an unsupervised or supervised learning method.

\subsection{Complexity}
\label{sec:complexity}

The initialization and post-processing steps in both algorithms are $O(|\mathcal{V}|^2)$ for a dense matrix implementation of the $W$s. However, they are sparse if $L << |\mathcal{V}|$. In this case, these steps can be computed in $O(L^2 \log{|\mathcal{V}|})$ or $O(L^2)$.

The processing step of Algorithm~\ref{alg:Parsing-Sequence-algo} has double nested iterations along the length of a sequence with unit computations within. Its complexity is $O(L^{2})$. On the other hand, the processsing in Algorithm~\ref{alg:Parsing-Sequence-algo-1} has double nesting along the alphabets also with unit computations therein. Therefore, a part of the computation is $O(|\mathcal{V}|^{2})$. Additionally, the helper function \verb|GetAlphabetPositions| is of order $O(|\mathcal{V}|L)$. Therefore, the processing complexity is $O(|\mathcal{V}|(L+|\mathcal{V}|))$. Thus, Algorithm~\ref{alg:Parsing-Sequence-algo-1} is more suitable if $L >> |\mathcal{V}|$.

The processing computation of SGT algorithms on a data set $\mathcal{S}$ with $n$ sequences will consequently be $O(nL^{2})$ and $O(n|\mathcal{V}|(L+|\mathcal{V}|))$ for Algorithm~\ref{alg:Parsing-Sequence-algo} and \ref{alg:Parsing-Sequence-algo-1}, respectively. However, the SGT embeddings of the sequences $s\in\mathcal{S}$ are independent. Note that the input to the algorithms is only one sequence $s$. Therefore, the embeddings can be computed in parallel for the sequences in a data set $\mathcal{S}$. It will be shown in \S~\ref{sec:parallel-computation-sequence-search} that parallel computation significantly reduces the runtime by distributing the sequences on several worker nodes and computing their embeddings simultaneously.

The size for a sequence embedding is $O(|\mathcal{V}|^2)$ in the worst case of no sparsity. In a worst case when every symbol $u\in\mathcal{V}$ is present in $s$, each element in the embedding will be non-zero. Otherwise, if the symbol set is large, typically only a fraction of them are present in $s$. In these cases, a sparse representation of the embedding reduces the size by a sparsity fraction $r$ to $O((r|\mathcal{V}|)^2), r\in(0,1)$. Therefore, although the size is quadratic with respect to $|\mathcal{V}|$ it is not an impediment.

Both the time complexity and size are independent of the tuning parameter $\kappa$. $\kappa$ adjusts the amount of short- and long-term dependencies captured in SGT embedding (refer to \S~\ref{sec:sgt-properties-short-long-term-features}). The computation complexities being independent of $\kappa$ gives a significant advantage to SGT. Unlike the existing methods, the short- and long-term dependencies to include can be tuned based on the problem and not restricted by computation limitations.

\subsection{Parameter Selection}
\label{sec:parameter-selection}

SGT embedding has only one tuning parameter $\kappa$. A small value of $\kappa$ allows longer-term dependencies captured in the embedding and vice-versa. In the implementation shown in this paper, $\kappa$ is chosen from $1,2,\ldots,10$. Although fractional values can also be taken but SGT's performance is insensitive to minor differences in $\kappa$. Therefore, $\kappa$ is chosen as integers in this paper.

The optimal selection of $\kappa$ depends
on the problem at hand. If the end objective is building a supervised
learning model, methods such as cross-validation can be used. For unsupervised
learning, any goodness-of-fit criteria can be used for the selection.
In cases of multiple parameter optimization, e.g. the number of clusters
(say, $n_{c}$) and $\kappa$ together in clustering,
we can use a random search procedure. In such a procedure, we randomly
initialize $n_{c}$, compute the best $\kappa$ based on some goodness-of-fit
measure, then fix $\kappa$ to find the best $n_{c}$, and repeat
until there is no change.

\section{Experimental Analysis} 
\label{sec:Experimental-Analysis}

SGT's efficacy can be assessed based on its ability to find
(dis)similarity between sequences. Therefore, 
we built a sequence clustering experimental setup. Clustering
requires an accurate computation of (dis)similarity between objects and
thus is a good choice for the efficacy test.

We show the following experiments here: a) Exp-1: length-sensitive sequence problem.
b) Exp-2: length-insensitive with \textbf{non-parametric} sequence pattern,
c) Exp-3: length-insensitive with \textbf{parametric} sequence pattern, and d) Exp-4: sensitivity analysis against the sample size, symbol set size, and sequence length.

The settings for each of them are given
in Table~\ref{tab:Experimentation-settings}. The table shows the mean and standard deviation of
the lengths of the sequences generated for each simulation in each experiment.
For Exp- 1, 2, and 4, a sequence is generated from some randomly
simulated set of motifs (strings) of random lengths (between 2-8).
These motifs are randomly placed and interspersed with arbitrary strings, which is the noise in a sequence. For experiments reproduction, details of sequence simulation are
in Appendix~\ref{sec:sequence-simulation}. In Exp-3, sequences are generated for a mixture of Markov and semi-Markov processes as presented in \citep{ferreira2005simulation}, and mixture of
Hidden Markov processes from \citep{helske2017mixture}.
In all the experiments, k-means clustering was applied to SGT representations of the sequences. Besides, the publicly available implementations of the benchmark methods were used.

\begin{table}[t]
\caption{Experimentation settings.}
\label{tab:Experimentation-settings}
\vskip -0.2in
\begin{center}
\begin{small}
\begin{sc}

\begin{tabular}{lccccc}
\hline 
\textbf{\scriptsize{}Experiment} & \textbf{\scriptsize{}Sequence length,$\mu,\sigma$} & \textbf{\scriptsize{}Sample size} & \textbf{\scriptsize{}Symbol set size} & \textbf{\scriptsize{}Noise} & \textbf{\scriptsize{}\#clusters,$n_{c}$}\tabularnewline
\hline 
{\scriptsize{}Exp-1} & {\scriptsize{}(424.6, 130.6)} & 500 & 26 & {\scriptsize{}45-50\%} & {\scriptsize{}5}\tabularnewline
{\scriptsize{}Exp-2} & {\scriptsize{}(116.4, 47.7)} & 500 & 26 & {\scriptsize{}35-65\%} & {\scriptsize{}5}\tabularnewline
{\scriptsize{}Exp-3} & {\scriptsize{}(98.2, 108.3)} & 500 & 26 & {\scriptsize{}\textendash{}} & {\scriptsize{}5}\tabularnewline
{\scriptsize{}Exp-4} & {\scriptsize{}{(100, -), (500, -), (1000, -)}} & {100, 500, 1000} & {20, 50, 100} & {\scriptsize{}35-65\%} & {\scriptsize{}5}\tabularnewline
\hline 
\end{tabular}
\end{sc}
\end{small}
\end{center}
\vskip -0.1in
\end{table}%


In Exp-1, we compared SGT with length-sensitive algorithms, viz. MUSCLE,
UCLUST, and CD-HIT, which are popular in Bioinformatics. These methods
are hierarchical in nature, and thus, they find the optimal number
of clusters. For SGT-clustering, the number of clusters is found using
the procedure recommended in \S\ref{sec:parameter-selection}. 

\begin{figure}
\centering
  \includegraphics[scale=0.4]{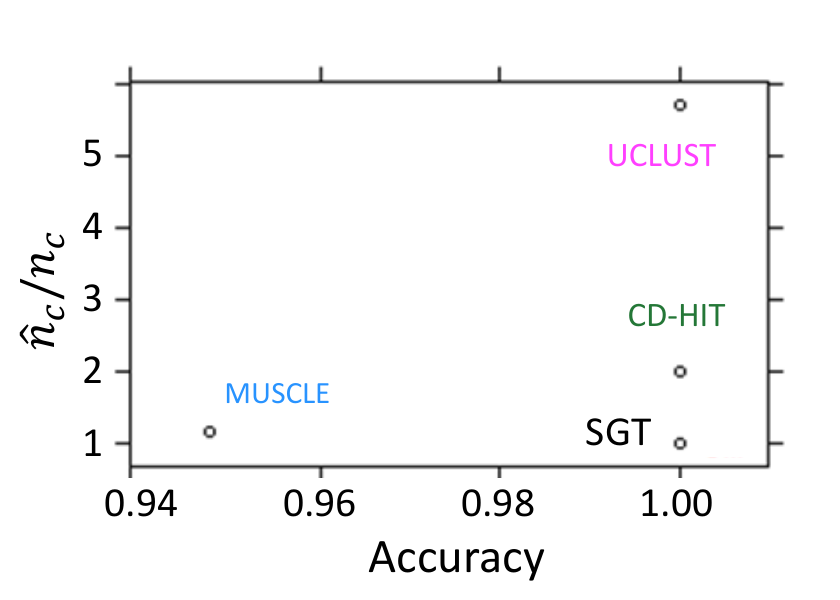}
  \caption{Exp-1 results.}
  \label{fig:Exp-1}
\end{figure}

Fig.~\ref{fig:Exp-1} shows the results, where the y-axis is the
ratio of the estimated best number of clusters, $\hat{n}_{c}$,
and the truth, $n_{c}$. The x-{}axis shows
the clustering accuracy.
For a best performing algorithm, both metrics should be close to 1.
As shown in the figure, CD-HIT and UCLUST overestimated the number of clusters
by about twice and five times, respectively. MUSCLE had a better $n_{c}$ estimate but
had about 95\% accuracy. On the other hand, SGT could accurately estimate
$n_{c}$ and has a 100\% clustering accuracy.

\begin{figure}
	\centering
    \subfloat[]{\label{fig:Overlap-f1}
    \includegraphics[scale=0.2]{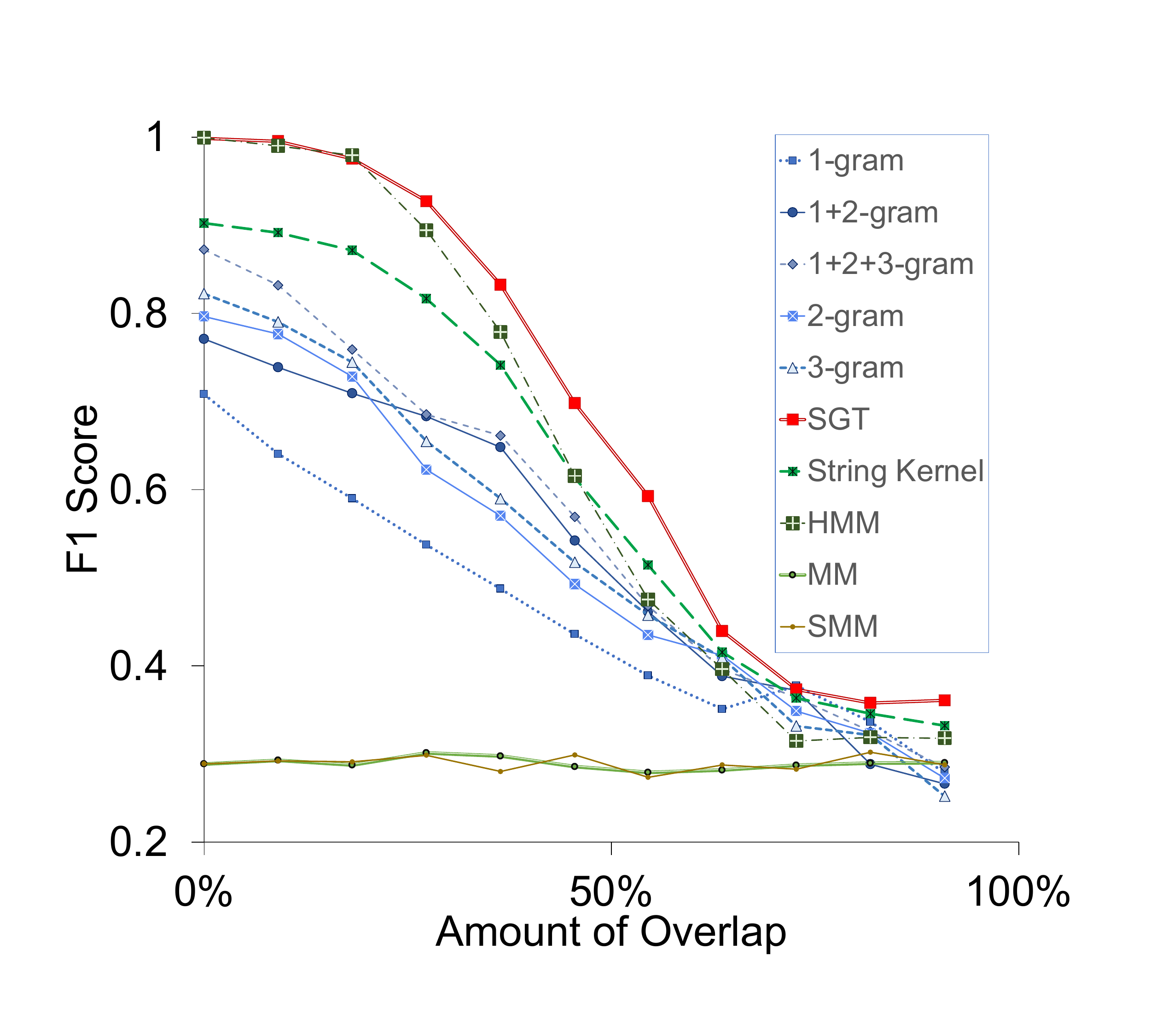}}
	\\
    \subfloat[]{\label{fig:Overlap-time}
    \includegraphics[scale=0.2]{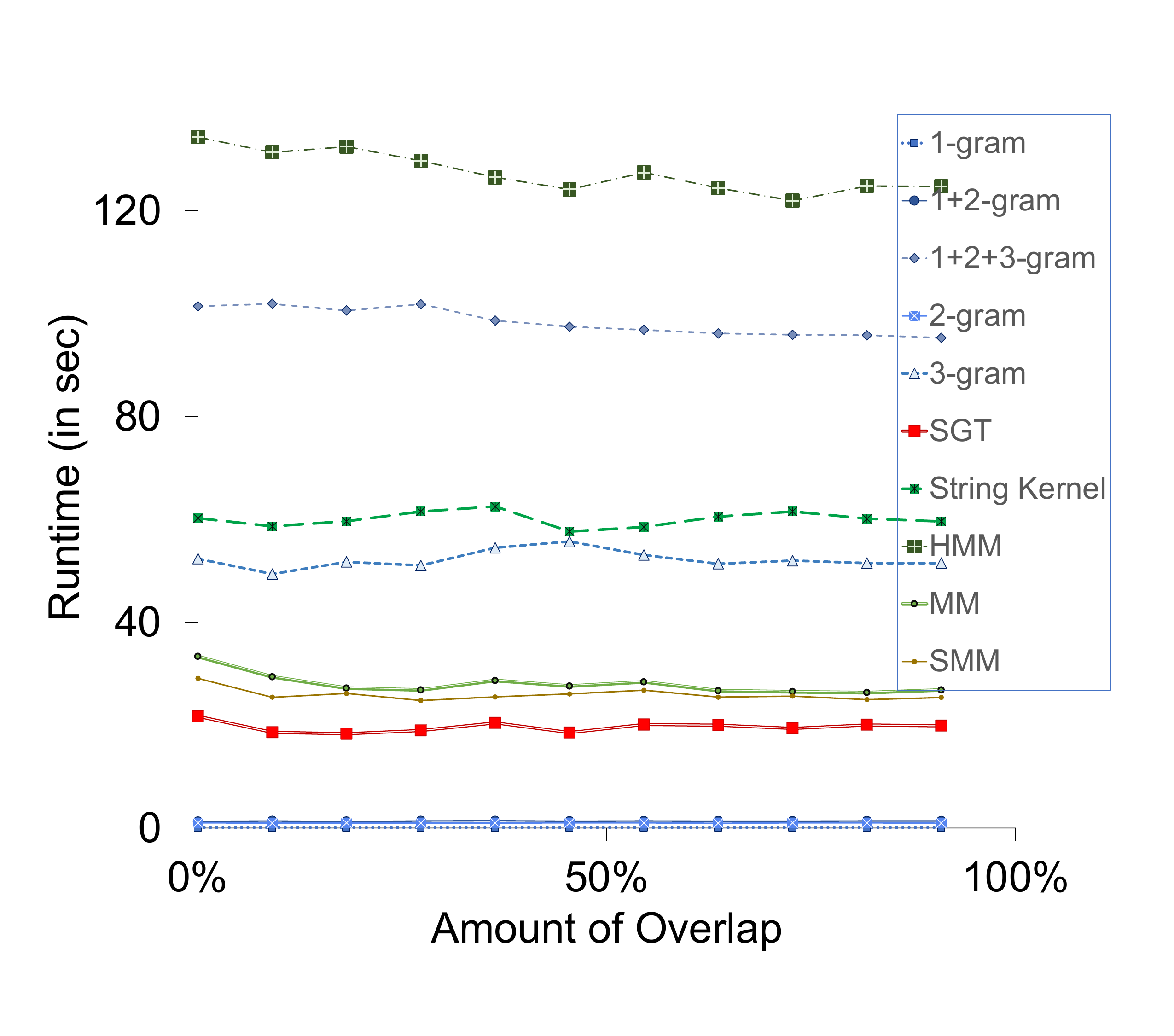}}
     
\caption{Exp-2 results.}
\end{figure}

In Exp-2, we compared SGT with popular and state-of-the-art 
sequence analysis
techniques, viz. n-gram, String Kernel, mixture Hidden Markov model (HMM), Markov
model (MM) and semi-Markov model (SMM)-based clustering. 
For n-gram,
we take $n=\{1,2,3\}$, and their combinations. In String Kernel, the subsequence length parameter $k=4$ is taken. For these methods, we provided
the known\emph{ }$n_{c}$ to the algorithms and report the F1-score for
accuracy. In this experiment, the clusters are increasingly overlapped
to make them difficult to separate. An overlapping cluster implies
their seed motifs set have a non-null intersection (see Appendix~\ref{sec:sequence-simulation}).

Exp-2's result in Fig.~\ref{fig:Overlap-f1} shows the accuracy (F1-score) and the runtimes in Fig.~\ref{fig:Overlap-time}, where SGT is seen to outperform all others in accuracy. MM and SMM have poorer accuracy because of the first-order Markovian assumption. HMM is found to have comparable accuracy, but its runtime is more than six times that of SGT. String Kernel and N-gram methods' accuracies lie in between. Low-order n-grams have smaller runtime than SGT but worse accuracy. Interestingly, the 1-gram method is better when overlapping
is high, showing the higher-order n-grams' inability to distinguish
between sequences when the overlap is high.

\begin{figure}
	\centering
    \subfloat[]{\label{fig:Beating-f1}
    \includegraphics[scale=0.2]{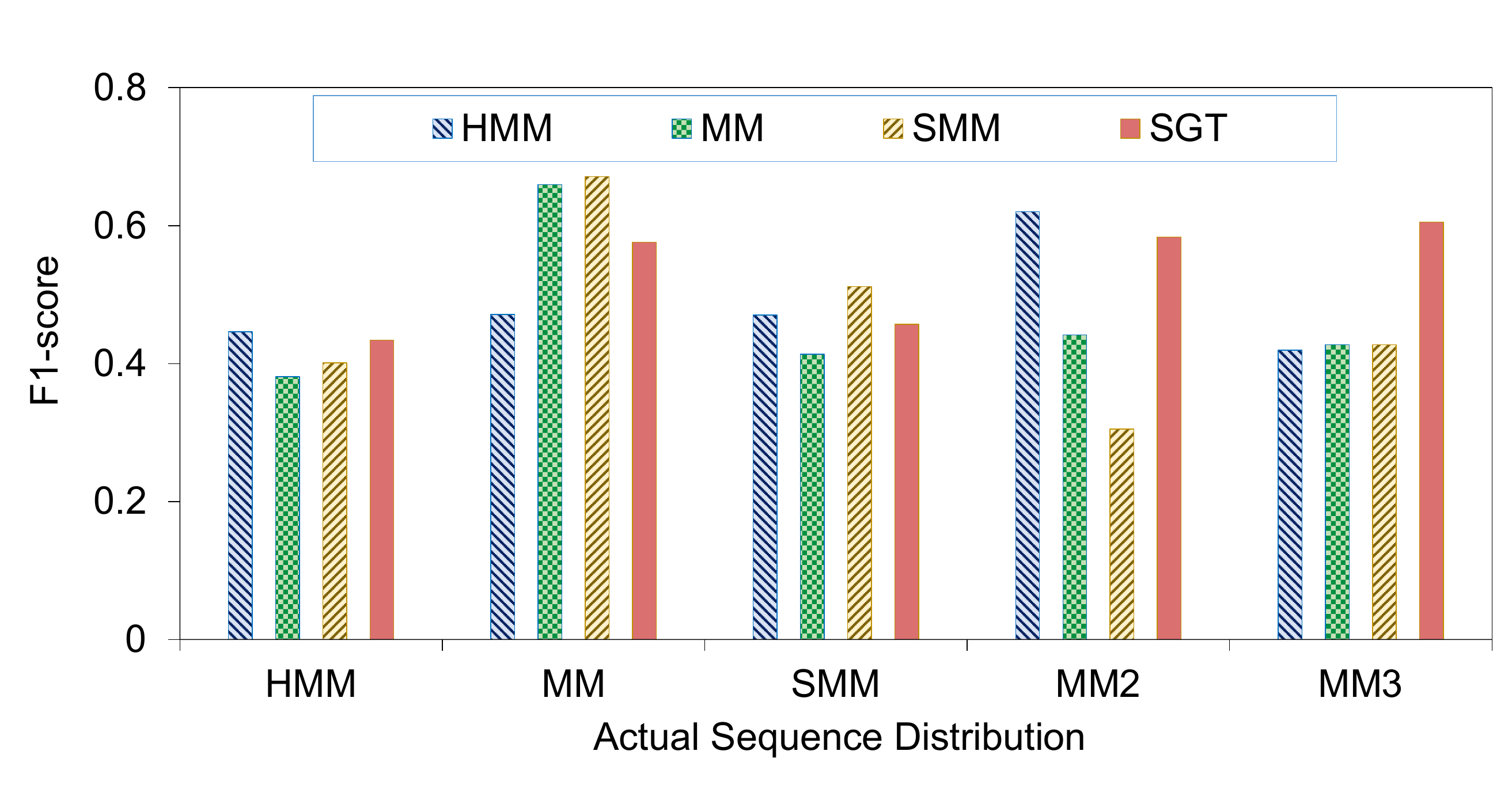}}
	\\
    \subfloat[]{\label{fig:Beating-time}
    \includegraphics[scale=0.2]{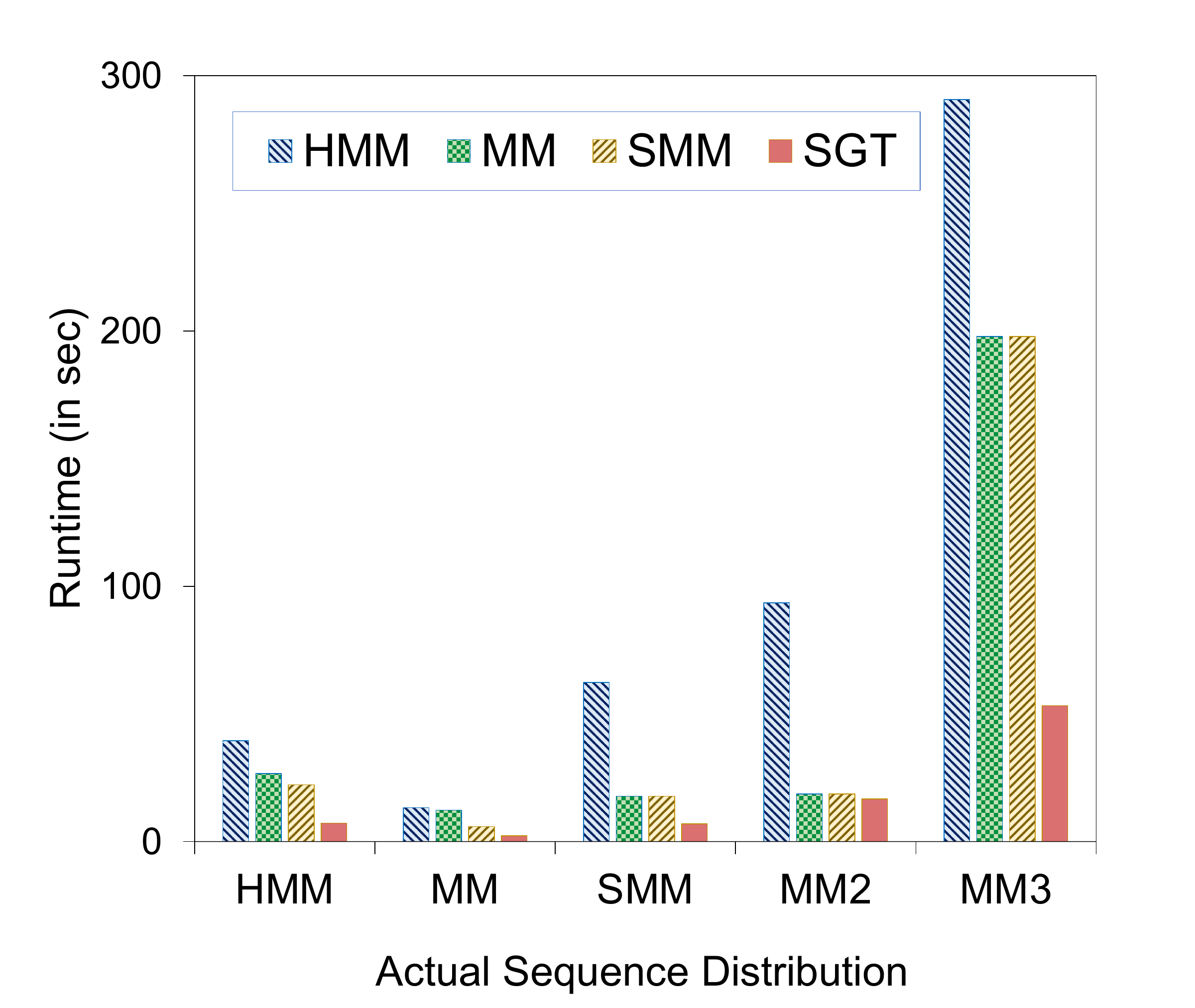}}
     
\caption{Exp-3 results.}
\end{figure}

Furthermore, we did Exp-3 to see the performance of SGT in sequence
data sets generated from the mixture of parametric distributions,
viz. the mixture of HMM, MM and SMM. The objective of this experiment
is to test SGT's efficacy on parametric data sets against parametric
methods. In addition to obtaining data sets from mixed HMM and first-order
mixed MM and SMM distributions, we also get second-order Markov (MM2)
and third-order Markov (MM3) data sets. Fig.~\ref{fig:Beating-f1} shows the F1-score and Fig.~\ref{fig:Beating-time} has the runtimes. As expected, the mixture clustering
method corresponding to the true underlying distribution is performing
the best. Note that SMM is slightly better than MM in the MM setting
because of its over-representative formulation, i.e. a higher dimensional
model to include a variable time distribution. However, the proposed
SGT's accuracy is always close to the best. This shows SGT's robustness
to underlying distribution and its universal applicability. And,
again, its runtime is smaller than all others.

\begin{figure}
	\centering
    \subfloat[]{\label{fig:sensitivity-sample-size-f1}
    \includegraphics[scale=0.35]{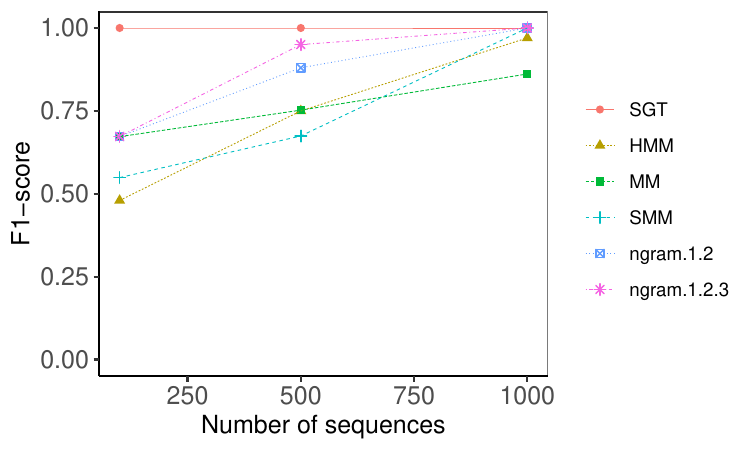}}
    \subfloat[]{\label{fig:sensitivity-sample-size-runtime}
    \includegraphics[scale=0.35]{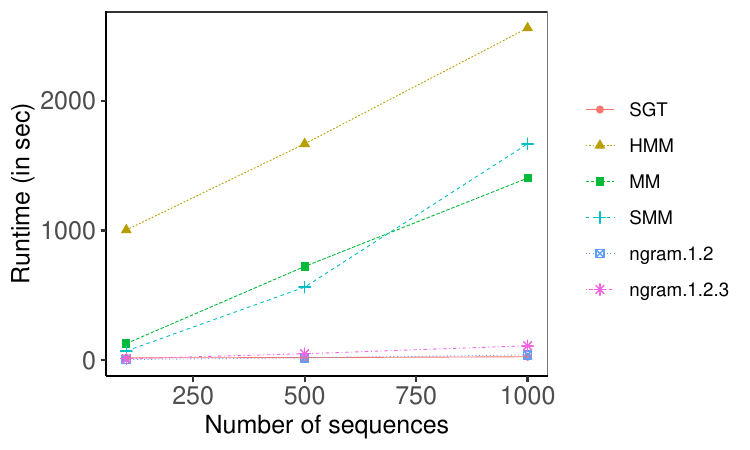}}
     
\caption{Exp-4 sensitivity analysis: sample size.}
\label{fig:exp-4-result-sample-size}
\end{figure}

\begin{figure}
	\centering
    \subfloat[]{\label{fig:sensitivity-symbols-runtime-f1}
    \includegraphics[scale=0.35]{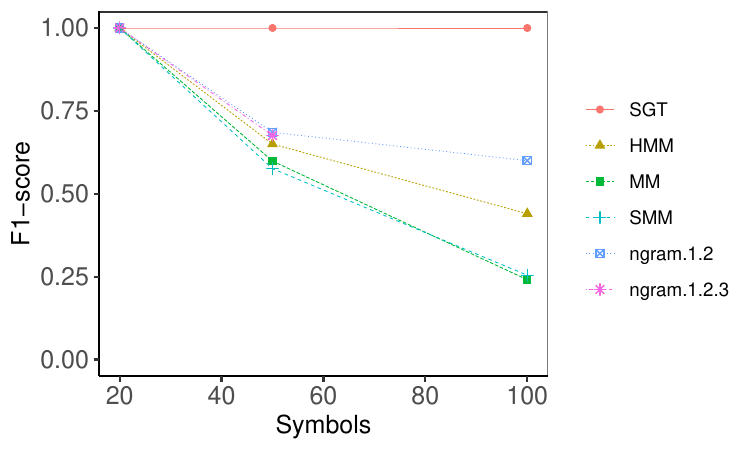}}
    \subfloat[]{\label{fig:sensitivity-symbols-runtime}
    \includegraphics[scale=0.35]{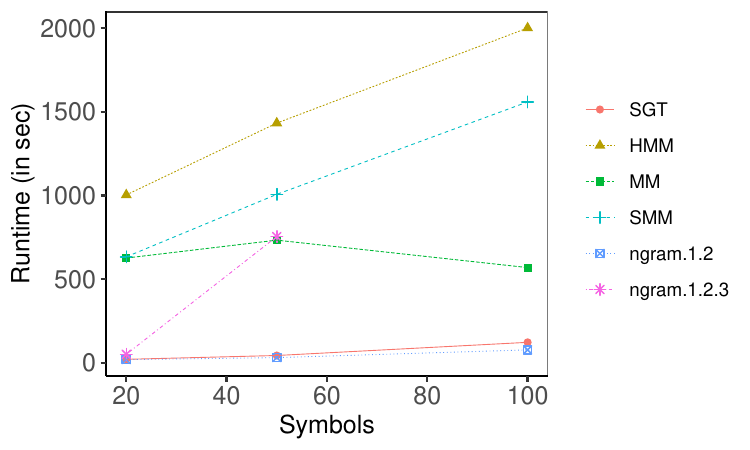}}
     
\caption{Exp-4 sensitivity analysis: symbol set size.}
\label{fig:exp-4-result-symbol-size}
\end{figure}

\begin{figure}
	\centering
    \subfloat[]{\label{fig:sensitivity-sequence-length-f1}
    \includegraphics[scale=0.35]{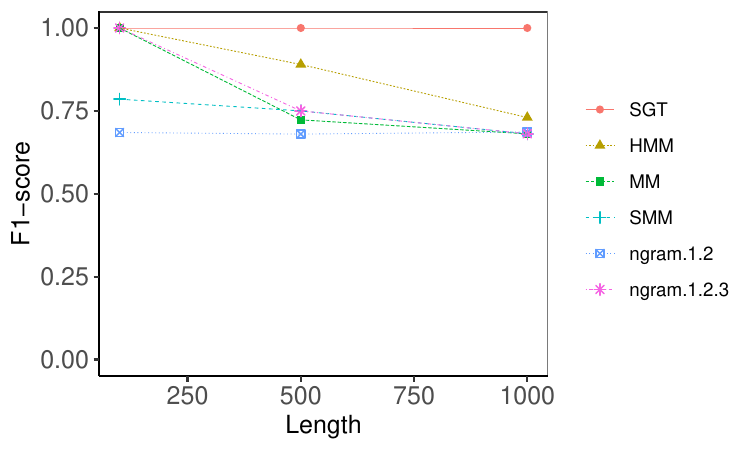}}
    \subfloat[]{\label{fig:sensitivity-sequence-length-runtime}
    \includegraphics[scale=0.35]{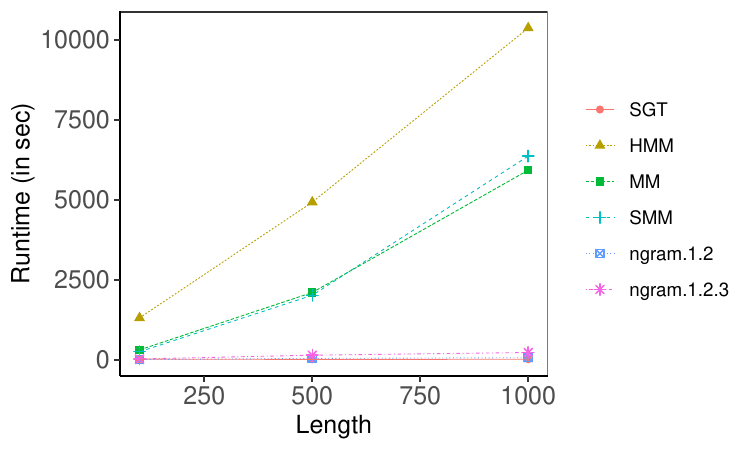}}
     
\caption{Exp-4 sensitivity analysis: sequence length.}
\label{fig:exp-4-result-sequence-length}
\end{figure}

Sensitivity analysis of SGT against the sample size, symbol set size, and sequence length is performed in Exp-4. As shown in Table~\ref{tab:Experimentation-settings}, the sample size, symbol set size, and the mean of sequence lengths ranged from $\{100, 500, 1000\}$, $\{20, 50, 100\}$, and $\{100, 500, 1000\}$, respectively. The standard deviation of the sequence lengths in this analysis was kept very small to accurately measure the effect of the length and, therefore, not mentioned in the table. Besides, the settings for the sensitivity analysis are not extreme values due to the computation limitation of some methods. For example, the 3-gram method failed to yield a result for a symbol set size of 100 on the test computing system.

The results are shown in Figure~\ref{fig:exp-4-result-sample-size}-\ref{fig:exp-4-result-sequence-length}. As shown in the charts for the f1-score, SGT embedding based k-means remained robust to the variations in sample size (Figure~\ref{fig:sensitivity-sample-size-f1}), symbol set size (Figure~\ref{fig:sensitivity-symbols-runtime-f1}), and sequence length (Figure~\ref{fig:sensitivity-sequence-length-f1}).

Among the benchmark methods, SMM is agnostic to the sequence length because it also more parameterization that incorporates the sequence length. Moreover, the non-parametric n-gram based methods were better than the parametric methods for varying symbol set size expect higher n-gram inability to yield results in tractable time for high feature space. 

Besides, while SGT's performance with the sample size was unaffected, the other methods improved with more samples. The runtime comparison shows that SGT embedding runtime although increases in each variation, is significantly lower than the benchmark methods. Non-parametric methods generally have lower runtime because the time-intensive feature estimation occurs only once. On the other hand, the iterative EM algorithm based estimation in the parametric methods re-computes the features at every iteration.

\section{Applications on Real Data} \label{sec:Applications}

The sequence mining problem can be broadly categorized as classification, clustering, and search. In the following, real-world examples for each of them are presented. Performance comparisons\footnote{Unless otherwise mentioned, the analyses are done on 2.2 GHz Quad-Core Intel Core i7 16 GB 1600 MHz DDR3 machine.} with state-of-the-art methods including deep learning is made on the labeled data in sequence classification. Sequence clustering demonstrated an application for unsupervised learning. Moreover, a sequence search is demonstrated using the parallel computation capability with SGT. The implementation steps are available at \url{https://github.com/cran2367/sgt}.

\subsection{Sequence Classification}
\label{subsec:classification}
Here we perform classification on a) protein
sequences\footnote{www.uniprot.org} having either of two known functions, which act as the labels, and b) network intrusion data\footnote{https://www.ll.mit.edu/ideval/data/1998data.html}
containing audit logs and any attack as a positive label. 

\begin{table}
\caption{data set attributes.\label{tab:data set-attributes}}
\begin{centering}
\begin{tabular}{l l l}
\hline 
\textbf{\scriptsize{}Attribute} & \textbf{\scriptsize{}Protein}{\scriptsize{}} & \textbf{\scriptsize{}Network}{\scriptsize{}}\tabularnewline
\hline 
{\scriptsize{}Sample size} & {\scriptsize{}2113} & {\scriptsize{}115}\tabularnewline
{\scriptsize{}Sequence length range} & {\scriptsize{}(289, 300)} & {\scriptsize{}(12, 1773)}\tabularnewline
{\scriptsize{}Class distribution} & {\scriptsize{}46.4\%+} & {\scriptsize{}11.3\%+}\tabularnewline
{\scriptsize{}Symbol set size} & {\scriptsize{}20 (amino acids)} & {\scriptsize{}49 (log events)}\tabularnewline
\hline 
\end{tabular}
\par\end{centering}
\end{table}

The data set details are in Table~\ref{tab:data set-attributes}. For both problems, we use the \emph{length-sensitive} SGT. For proteins, it is due to their nature, while for network logs, the lengths are important because sequences with similar patterns
but different lengths can have different labels. Consider a simple example of two sessions: \{\texttt{\small{}login, pswd, login,
pswd,...}\} and \{\texttt{\small{}login, pswd,...(repeated
several times)..., login, pswd}\}. While the first session can
be a regular user mistyping the password once, the other session is
possibly an attack to guess the password. Thus, the sequence lengths
are as important as the patterns.

For the network intrusion data, the sparsity of SGTs was high. Therefore,
we performed principal component analysis (PCA) on it and kept the
top 10 PCs as sequence features. We call it SGT-PC, for further modeling. For proteins, the SGTs are used directly. SVM classifier is trained
on \textit{n}-grams with an RBF kernel, cost parameter set 
to 1, and on current state-of-the-art String Kernels \citep{kuksa2009scalable}, its faster approximate improvement by \citep{farhan2017efficient} (the \textit{k}-mer and \textit{mismatch} lengths are set to 5 and 2, respectively), and a recently developed state-of-the-art Random Features Kernel\citep{wu2019efficient}. Table~\ref{tab:Classification-accuracy-(F1-score)} reports the
average test accuracy (F1-score) from a 10- and 5-fold cross-validation
for the protein and network data, respectively.

\begin{table}
\caption{Classification accuracy (F1-score)
	results.\label{tab:Classification-accuracy-(F1-score)}}
\noindent \begin{centering}
{\scriptsize{}}%
\begin{tabular}{l l l}
\hline 
\textbf{\scriptsize{}SVM on- \{$\gamma_{protein};\gamma_{network}$\}} & \textbf{\scriptsize{}Protein } & \textbf{\scriptsize{}Network}\tabularnewline
\hline 
{\scriptsize{}SGT\{$0.0014;0.1$\}} & \textbf{\scriptsize{}99.61\%,\(\kappa=1\)} & \textbf{\scriptsize{}89.65\%,\(\kappa=10\)}\tabularnewline

{\scriptsize{}2-gram\{$0.0025;0.00041$\}} & {\scriptsize{}93.87\%} & {\scriptsize{}63.12\%}\tabularnewline
{\scriptsize{}3-gram\{$0.00012;8.4e-6$\}} & {\scriptsize{}95.12\%} & {\scriptsize{}49.09\%}\tabularnewline
{\scriptsize{}1+2-gram\{$0.0012;0.0004$\}} & {\scriptsize{}94.34\%} & {\scriptsize{}64.39\%}\tabularnewline
{\scriptsize{}1+2+3-gram\{$4.0e-5;8.2e-6$\}} & {\scriptsize{}96.89\%} & {\scriptsize{}49.74\%}\tabularnewline
\hline 
{\scriptsize{}String Kernel \citep{kuksa2009scalable}} & {\scriptsize{}97.63\%} & {\scriptsize{}52.36\%}\tabularnewline
{\scriptsize{}Fast Kernel \citep{farhan2017efficient}} & {\scriptsize{}94.66\%} & {\scriptsize{}41.76\%}\tabularnewline
{\scriptsize{}Random Features Kernel \citep{wu2019efficient}} & {\scriptsize{}96.18\%} & {\scriptsize{}55.24\%}\tabularnewline
\hline 
\end{tabular}
\par\end{centering}{\scriptsize \par}
\end{table}

\begin{table}
\caption{Deep Learning\label{tab:dl-protein}}
\begin{centering}
\begin{minipage}[b][1\totalheight][t]{1.0\columnwidth}%
\begin{center}
\subfloat[Protein data.\label{tab:dl-protein}]{
\begin{tabular}{l l l}
\hline 
\textbf{\scriptsize{}Setting} & \textbf{\scriptsize{}F1-Score}{\scriptsize{}} & \textbf{\scriptsize{}Runtime}{\scriptsize{}}\tabularnewline
\hline 
{\scriptsize{}LSTM-1 hidden layer- LSTM(32)} & {\scriptsize{}0.997} & {\scriptsize{}272  sec}\tabularnewline
{\scriptsize{}LSTM-2 hidden layer- LSTM(32),LSTM(16)} & {\scriptsize{}1.000} & {\scriptsize{}415 sec}\tabularnewline
{\scriptsize{}FNN-1 hidden layer- Dense,Relu(16)} & {\scriptsize{}1.000} & {\scriptsize{}10 sec}\tabularnewline
{\scriptsize{}FNN-2 hidden layer- Dense,Relu(32),Relu(16)} & {\scriptsize{}1.000} & {\scriptsize{}11 sec}\tabularnewline
\hline 
\end{tabular}
}
\end{center}
\end{minipage}

\begin{minipage}[b][1\totalheight][t]{1.0\columnwidth}%
\begin{center}
\subfloat[Network data.\label{tab:dl-network}]{
\begin{tabular}{l l l}
\hline 
\textbf{\scriptsize{}Setting} & \textbf{\scriptsize{}F1-Score}{\scriptsize{}} & \textbf{\scriptsize{}Runtime}{\scriptsize{}}\tabularnewline
\hline 
{\scriptsize{}LSTM-1 hidden layer- LSTM(128)} & {\scriptsize{}0.353} & {\scriptsize{}467 sec}\tabularnewline
{\scriptsize{}LSTM-2 hidden layer- LSTM(64),LSTM(32)} & {\scriptsize{}0.502} & {\scriptsize{}822 sec}\tabularnewline
{\scriptsize{}FNN-1 hidden layer- Dense,Relu(128)} & {\scriptsize{}0.663} & {\scriptsize{}23 sec}\tabularnewline
{\scriptsize{}FNN-2 hidden layer- Dense,Relu(64),Relu(32)} & {\scriptsize{}0.563} & {\scriptsize{22 sec}}\tabularnewline
\hline 
\end{tabular}
}
\end{center}
\end{minipage}

\end{centering}
\end{table}

As we can see in Table~\ref{tab:Classification-accuracy-(F1-score)},
the F1-scores are high for all methods in the protein data, with SGT-based SVM surpassing all others. On the other
hand, the accuracies are small for the network intrusion data. This
is primarily due to a) a small data set but high dimension (related
to the symbol set size), leading to a weak predictive ability of models,
and b) a few positive class examples (unbalanced data) causing a poor
recall rate. Still, SGT outperforms other methods by a significant
margin. 

Furthermore, LSTM models in Deep Learning are state-of-the-art
for sequence classification. We compare it with a regular FNN Deep Learning
models (an MLP, specifically) in which the SGT features are used as
embedding layer. For learning, we use \textit{binary cross-entropy} loss function and \textit{adam} optimizer. Tensorflow is used for the implementations.

Table~\ref{tab:dl-protein}-\ref{tab:dl-network} shows the
results. The accuracies (F1-scores) in the
Protein data is close to 1 for all models. SGT powered FNN is only
marginally better in accuracy, however, its runtime is a fraction of
LSTM's runtime. In the network data, the accuracies are lower for LSTM models. This can be because, a) it is a length
sensitive sequence problem, b) the sequence lengths vary significantly.
LSTMs may not capture the differences due to lengths. Also, LSTMs pad
all sequences to become equal lengths, which may not work as effectively if
the differences in lengths are significantly high (LSTMs still performed well on protein data because the difference in the lengths is quite small). On the other hand, the FNN worked reasonably better. Its accuracy is higher than SVM on the other methods but smaller than SVM on SGT. This can be due to a small data set which makes the model training more difficult for a deep learning model. For the same reason, a single layer FNN worked better than two-layer.

\subsection{Sequence Clustering\label{sec:Example:-Understanding-user}}

We perform clustering user activity
on the web (weblog sequences) to understand user behavior.

We took users' navigation data (weblogs)
on msnbc.com \footnote{http://archive.ics.uci.edu/ml/data sets/msnbc.com+anonymous+web+data} collected during a 24-hour
period. The symbols of these sequences are the events corresponding
to a user's page request, e.g. \texttt{frontpage, tech}, etc. There are a total of 12 types of events ($|\mathcal{V}|$ = 12). The data set has a random sample of
100,000 sequences. The sequences' average length and standard deviation is (6.9, 27.3), with the range between (2, 7440) and skewed distribution.

\begin{figure}
    \centering
    \includegraphics[scale=0.4]{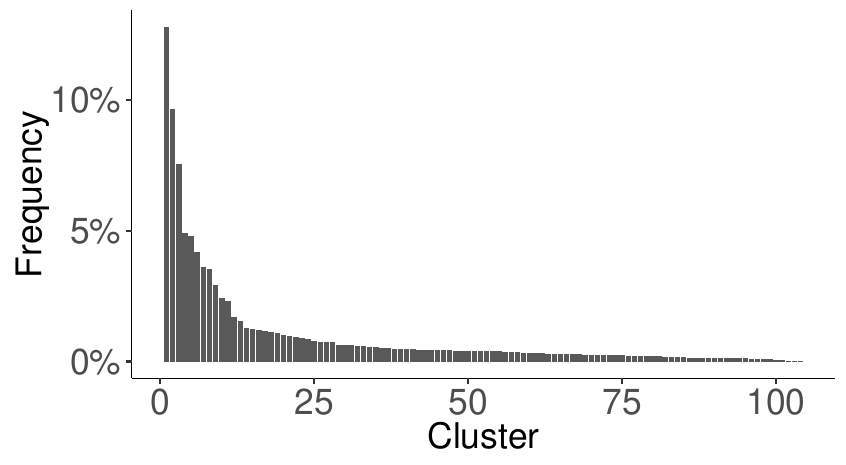}
    \caption{Clustering.}
    \label{fig:Frequency-distribution-of-msnbc-clusters}
\end{figure}

Our objective is to cluster the users with similar navigation patterns,
irrespective of differences in their session lengths.
We, therefore, take the \emph{length-insensitive }SGT and
use the random search procedure for optimal clustering in \S\ref{sec:parameter-selection}. We performed k-means clustering and the goodness-of-fit
criterion as db-index and found the optimality for $\kappa=9$ at $n_{c}=104$, which is close to the result in \citep{cadez2003model}. The frequency distribution (Fig.~\ref{fig:Frequency-distribution-of-msnbc-clusters}) of the number of members in each cluster
 has a long-tail\textemdash{}the majority of users belong to a small set of clusters.


\begin{figure}
	\centering
    \subfloat[Cluster \#1]{\label{fig:Cluster=000023-1}
    \includegraphics[scale=0.15]{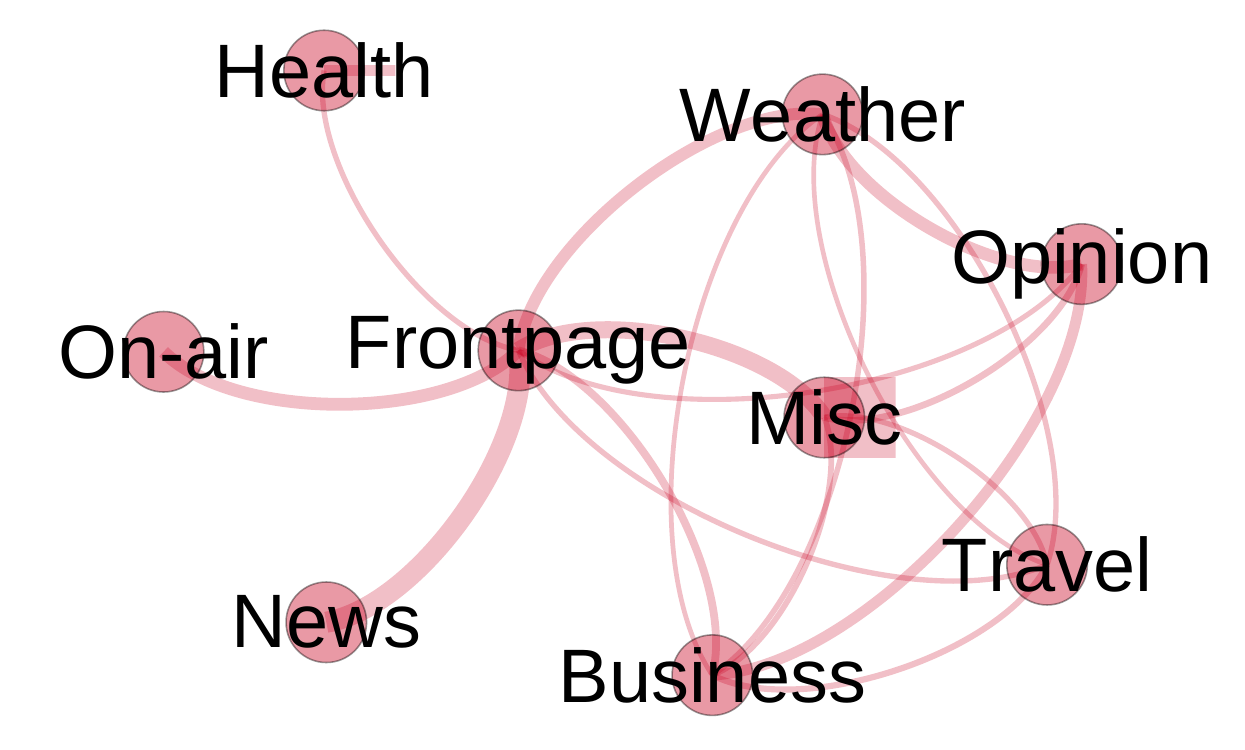}}
    \subfloat[Cluster \#3]{\label{fig:Cluster=000023-3}
    \includegraphics[scale=0.10]{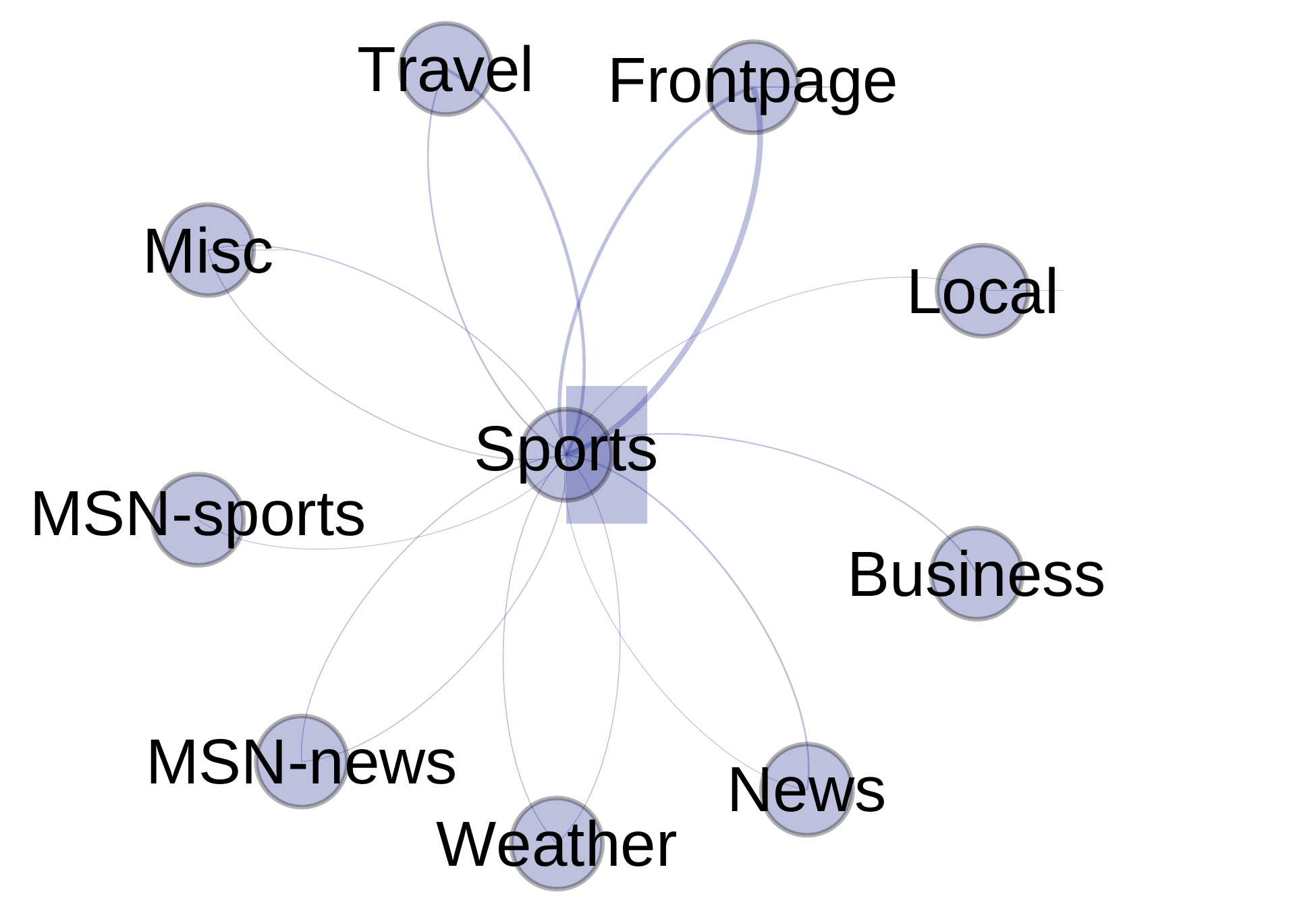}}
     
\caption{Graphical visualization of cluster centroids.}
\end{figure}






Additionally, SGT enables a visual interpretation of the clusters. In Fig.~\ref{fig:Cluster=000023-1}-\ref{fig:Cluster=000023-3},
we show a graph visualization of some clusters' centroids (which are in the SGT space), a representative of a behavior.

\subsection{Parallel Computation and Sequence Search}
\label{sec:parallel-computation-sequence-search}

\begin{table}
\caption{Parallel computing cluster specifications.\label{tab:parallel-computing-cluster-specifications}}
\noindent
\begin{centering}

\begin{tabular}{ll}
\hline
Parallel Computing Cluster & Specification                  \\ \hline
Driver Type                & r4.xlarge                      \\
                           & 30.5 GB Memory, 4 Cores, 1 DBU \\
Worker Type                & r4.xlarge                      \\
                           & 30.5 GB Memory, 4 Cores, 1 DBU \\
                           & Min Workers: 2                 \\
                           & Max Workers: 24                \\ \hline
\end{tabular}
\end{centering}
\end{table}

Typically sequence databases found in the real world are quite large. For example,
protein databases have millions of sequences and increasing. Here
we show that SGT sequence features can lead to a fast and accurate
sequence search. More specifically, we will utilize parallel computation capability possible with SGT.

We collected a sample of 10k and 1M protein sequences from the UniProtKB database on \url{www.uniprot.org}. First, we ran a benchmark test for runtime comparison. We computed the SGT embeddings for the two data set with the default mode and parallel computation mode. In the default mode, the embeddings are computed one-by-one. Therefore, the computation will be proportional to the sample size. 

In the parallel computation mode, a data set is partitioned into smaller chunks and distributed over several worker nodes. The embeddings are computed in parallel on these worker nodes. Depending on the number of worker nodes and their capacity, the data set can be repartitioned to more chunks and the overall runtime can be reduced.

The specifications of the parallel computing cluster are shown in Table~\ref{tab:parallel-computing-cluster-specifications}. The cluster was hosted on AWS. The driver and worker nodes were \verb|r4.xlarge|\footnote{The configurations details of \texttt{r4.xlarge} is available  here \url{https://docs.aws.amazon.com/AWSEC2/latest/UserGuide/memory-optimized-instances.html}.}.

The runtimes for SGT embedding computation are presented in Table~\ref{tab:parallel-computation-runtime}. As shown in the table, the runtime for 10k data set reduced from 13.5 minutes under the default mode to about 30 seconds with parallel computation. In this run, the data set was partitioned into 500 chunks. The runtime reduction is more significant for the 1 Million data set. In this case, the data set was partitioned into 10k chunks and the runtime reduced from more than 24 hours to less than 30 minutes.

\begin{table}[]
\caption{Parallel computation runtime comparison.\label{tab:parallel-computation-runtime}}
\noindent
\begin{centering}
\begin{tabular}{lll}
\hline
\textbf{Sample size} & \textbf{Default mode runtime} & \textbf{Parallel computation runtime (repartitions)} \\ \hline
10k                  & 13.50 minutes            & 31.2 seconds (500)                        \\
1 Million            & 24+ hours                & 28.96 minutes  (10k)                         \\ \hline
\end{tabular}
\end{centering}
\end{table}

\begin{table}
\caption{Protein search query (A0A2T0PYE0) result.\label{tab:Protein-search-query-result}}
\noindent \begin{centering}
{\footnotesize{}}%

\begin{tabular}{lll}
\hline
\textbf{SGT} & \textbf{BLAST} & \textbf{CLUSTAL-Omega} \\ \hline
K0ZGN5       & V5XH98         & V5XH98                 \\
A0A0Y1CPH7   & A0A545TVA1     & K0K3L7                 \\
A0A5R8LCJ1   & K0K3L7         & A0A545TVA1             \\
K0K3L7       & A0A1I1B4Y2     & A0A3S0HQH6             \\
A0A0N9I8D1   & A0A0N9I8D1     & A0A0N9I8D1             \\ \hline
\end{tabular}
\par\end{centering}{\footnotesize \par}
\end{table}

The embeddings are then stored and a sequence query search is performed. A protein sequence (id: A0A2T0PYE0) is arbitrarily chosen. The objective is to find protein sequences similar to A0A2T0PYE0 in the data set.

At this stage, the embeddings of the database sequences are known. To search sequences similar to the query we compute the embedding for the query A0A2T0PYE0\footnote{The protein sequence of A0A2T0PYE0 is,\\ \texttt{MSAAADRPTVEISTDFYSLDALMALVDEPPRLALAPEVAERIDAGARYVERIAPQDRHIY\\
GINTGFGPLCETRIPADRMSELQHKHLVSHACGVGEPVPERVSRLAMLVKLLTFRAGYSG\\
ISLEAVQRVLDLWNADVIPVVPKKGTVGASGDLAPLAHLALPLIGLGKVRVDGRITDAGA\\
VLEAMGWKPLRLKPKEGLALTNGVQYINALALDSVLRSERLIKAADLIAGLSIQGFSCAD\\
TFYQPILHATSLHPERSAVAGNLVRLLDGSNHHTLPQGNAAREDPYSFRCAPQVHAAVRQ\\
TCGFARDIVGRECNSVSDNPLFFPEHDQVILGGNLHGESTAFALDFLAIAMSELANISER\\
RTYQLLSGQHGLPDFLAPEPGVDSGLMIPQYTSAALVNENKVLATPASIDTIPTSQLQED\\
HVSMGGTSAYKLWTILDNCEYVLAVELMTAVQAIDLNQGLRPSPATRGVVAEFRQEVGFL\\
REDRLQADDIEKSRRYLRGRLRTWAKDLD}}. The dot product of the query embedding with the embeddings in the database is computed. A dot product is a measure of the similarity. Among the various choices of similarity measure, the dot product is chosen here because it is computationally faster with embeddings. 

The sequences with large dot products will have high similarity with the query. The top five similar protein sequences from the data set are shown in Table~\ref{tab:Protein-search-query-result}. The top five for the commonly used protein search methods BLAST and CLUSTAL-Omega are also shown for reference. 

Only two of the five SGT search results match with BLAST and CLUSTAL-Omega while the result of the latter two are more similar. This is expected because both the latter methods are alignment based. On the other hand, SGT looks for similarity in the distribution of sequence symbol positions. Such a distribution similarity based search is quite applicable on user data, e.g., for behavioral analysis as presented in \S~\ref{sec:Example:-Understanding-user}\footnote{Essentially, SGT-based search could be used for problems where distribution-based methods like Markov or Hidden Markov model are used as opposed to alignment.}. 

The search operation can be further improved with respect to accuracy and speed by applying a dimension reduction using methods like PCA and performing the dot product on the reduced dimension.

\section{Discussion: Why does SGT work better?}

\subsection{Ability to work in length- sensitive and insensitive problems}
\label{subsec:ability-to-work-in-length}

\begin{figure}
\centering

  \includegraphics[scale=0.52]{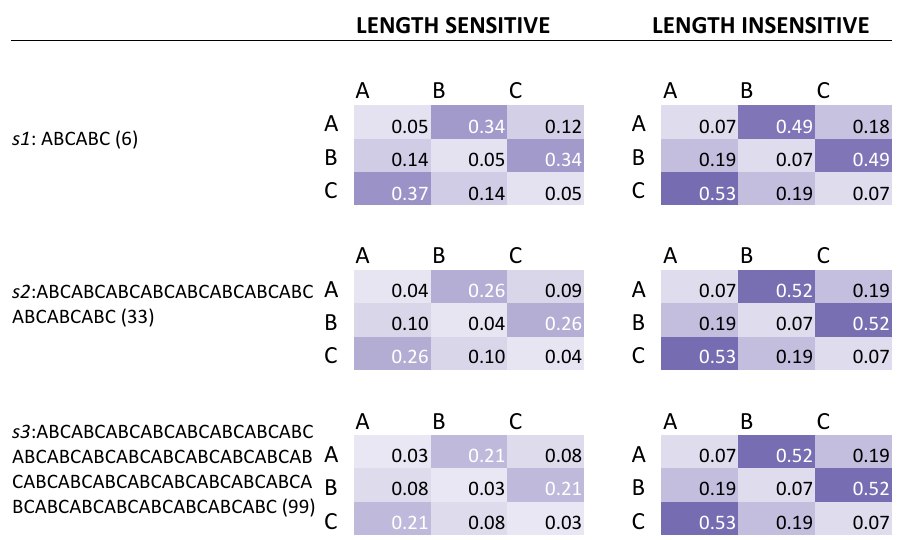}
  \caption{Effect of sequence length on SGT features ($\kappa = 5$).}
  \label{fig:why-sgt-works-better}

\end{figure}

We discussed SGT's length- sensitive and insensitive variants in
\S~\ref{subsec:SGT-definition} and \ref{sec:sgt-properties-short-long-term-features}.
Here we show SGT's ability to work in both with a real example.
Consider the three sequences, $s1$, $s2$, and $s3$,
in Fig.~\ref{fig:why-sgt-works-better} The sequences are of lengths
6, 33, and 99, respectively. Their inherent pattern is $\{\mathtt{A},
\mathtt{B},\mathtt{C}\}$ occurring in succession.

We find their SGT features for both length- sensitive and insensitive
variants. We show the SGT features as adjacency matrices in 
Fig.~\ref{fig:why-sgt-works-better}. We will first look at the
length insensitive column. We notice that the SGT features $s1$
and $s2$ are quite close. And as the sequence lengths increased
the SGT features approached a constant value. This can be noted
from the length-insensitive SGT features of $s2$ and $s3$ which
are the same up to two decimals.

On the other hand, in the length-sensitive case, the SGT features
keep changing as the length changes. Note that the features
reduce consistently as the length increases. This is because,
as shown in Theorem~\ref{theorem:closed-form}, the expectation
of the length-sensitive SGT features reduce with length. To avoid the features from approaching zero
(for high lengths), we can tune the hyperparameter $\kappa$.

This example reinstates that SGT can effectively take into
account the length- sensitivity and insensitivity. Moreover, the features derived from the SGT algorithm in Fig.~\ref{fig:why-sgt-works-better}	are approximately equal to the value computed from Equation~\ref{eq:E-phi-AB-part-2-2} confirming the theoretical interpretations in \S~\ref{sec:sgt-properties-short-long-term-features}.

The sequences
in this example were noise-free. However, SGT is robust to
noise which we will discuss in \S~\ref{subsec:robust-to-noise}.

\subsection{Avoids false positives by inherently accounting for mismatches}

In this paper, a false positive is defined as identifying two
sequences of different lengths as similar if the smaller sequence
is a subsequence of or locally aligns with the longer sequence.
Avoiding false positives is a nontrivial challenge in 
length-insensitive sequence problems. For example,
\textit{N}-gram methods can often lead to such false positives.
To address this, mismatch Kernels were developed 
\citep{eskin2003mismatch}. However, these methods require
additional computations for the mismatches or substitutions. 
On the other hand, SGT inherently accounts for the mismatches.

Consider a small sequence $\mathtt{ABCABC}$ and compare it with
$\mathtt{ABCABCABCABCABCABC}$ and $\mathtt{ABCABC\textcolor{red}{DEFGHIJKLMNO}}$. Ideally, we require $\mathtt{ABCABC}$ to match
with the former but not the latter. As shown in 
Table~\ref{tab:false-positive-table}, SGT feature comparisons
achieve this and, thus, avoids a false positive.

\begin{table}
\caption{SGT accounting for mismatches ($\kappa = 5$).\label{tab:false-positive-table}}
\noindent \begin{centering}
{\footnotesize{}}%
\begin{tabular}{lc}
\hline 
\textbf{\scriptsize{}Sequences comparison} & \textbf{\scriptsize{} Norm-1 difference }\tabularnewline
\hline 
{\small{}$\mathtt{ABCABC}$ vs $\mathtt{ABCABCABCABCABCABC}$} & {\small{}0.001}\tabularnewline
{\small{}$\mathtt{ABCABC}$ vs $\mathtt{ABCABC\textcolor{red}{DEFGHIJKLMNO}}$} & {\small{}5.880}\tabularnewline
\hline 
\end{tabular}
\par\end{centering}{\footnotesize \par}
\end{table}

\subsection{Robust to noise}
\label{subsec:robust-to-noise}
\begin{figure}
\begin{centering}
\includegraphics[scale=0.2]{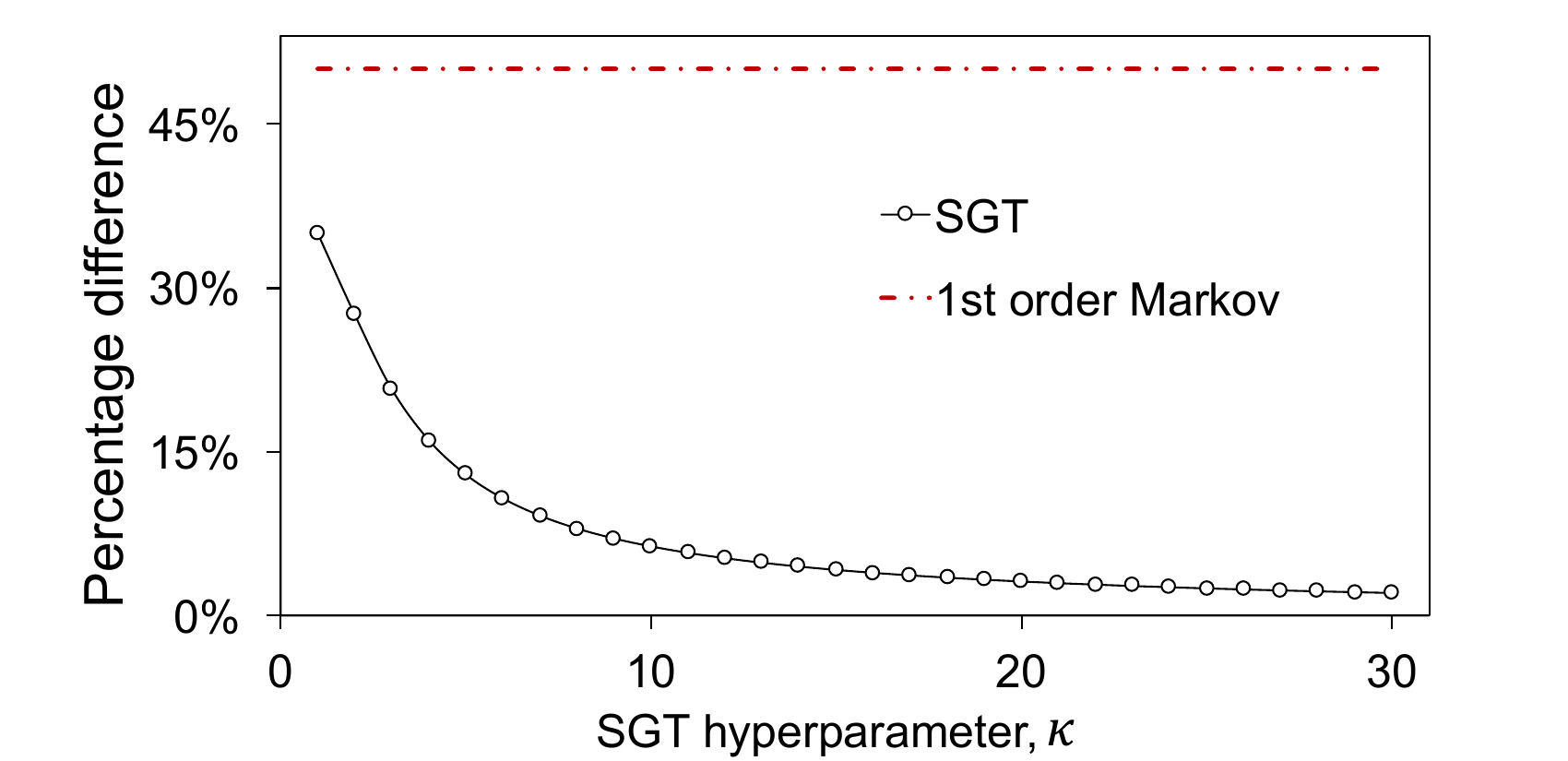}
\par\end{centering}
\caption{Percentage change in SGT feature for ($\mathtt{A}$,$\mathtt{B}$)
with $\kappa$ in the presence of noise.\label{fig:Percentage-change-in-SGT-feature}}
\end{figure}

To explain SGT's robustness to noise we will draw parallels 
with Markov Model. Compare SGT with a first-order Markov model.  
Suppose we are analyzing sequences in which ``$\mathtt{B}$
occurs \emph{closely} after $\mathtt{A}$.'' Due to stochasticity, the observed sequences can be like: a) \textbf{$\mathtt{\mathbf{AB}CDE\mathbf{AB}}$}, and b) $\mathtt{\mathbf{AB}CDE\mathbf{A}X\mathbf{B}}$, where (b) is same as (a) but a noise $\mathtt{X}$, appearing in between
$\mathtt{A}$ and $\mathtt{B}$. While their transition probabilities $P(A \rightarrow B)$ in sequence (a) and (b) in a Markov model is significantly different (a:1.00 and b:0.50), SGT is robust to such noises. The SGT feature for ($\mathtt{A}$,$\mathtt{B}$) for length- sensitive and insensitive scenarios are (a:0.50 and b:0.45), and (a:0.34 and b:0.30), respectively, for $\kappa=5$. 
As shown in Fig.\ref{fig:Percentage-change-in-SGT-feature},
the percentage change in the SGT feature for ($\mathtt{A}$,$\mathtt{B}$),
in the above case, is smaller than the Markov and decreases with increasing
$\kappa$. It also shows that we can easily regulate the effect of
such stochasticity by changing $\kappa$: choose a high $\kappa$ to reduce the noise effect, 
with a caution that sometimes the interspersed
symbols may not be noise but part of the sequence's pattern (thus,
we should not set $\kappa$ as a high value without a validation).
Furthermore, a Markov model cannot easily distinguish between these
two sequences: \textbf{$\mathtt{\mathbf{AB}CDE\mathbf{AB}}$} and
$\mathtt{\mathbf{AB}CDEFGHIJ\mathbf{AB}}$, from the ($\mathtt{A}$,$\mathtt{B}$)
transition probabilities (=1 for both). Differently, the SGT feature for ($\mathtt{A}$,$\mathtt{B}$)
changes from 1.72 to 2.94 ($\kappa=1$), because it looks at the overall pattern. 

\section{SGT Extensions}
\label{sec:Extensions-of-SGT}

\subsection{SGT for Bidirectional Sequences}
\label{subsec:sgt-for-undirected-sequences}

As also mentioned in \S\ref{subsec:SGT-Overview}, in a \textit{bidirectional}
sequence the chronological order of events does not matter, e.g. proteins.
This is also the case with weblog data, such as
music listening history, where sometimes we want to understand the 
tracks that were listened together and not the order in which they were
played.

SGT for bidirectional sequences can be computed by just changing the
definition of $\Lambda_{uv}(s)$ in Eq.~\ref{eq:psi-lambda-main} to,

\begin{eqnarray}
\tilde{\Lambda}_{uv}(s) & = & \{(l,m):\,s_{l}=u,s_{m}=v,\nonumber \\
 &  & l,m\in1,\ldots,L^{(s)}\}
 \label{eq:psi-lambda-undirected}
\end{eqnarray}

Below we show that under the assumption---all the symbols are
present in a sequence with a uniform probability---, the bidirectional SGT features can be approximated directly from the directed.

We can write Eq.~\ref{eq:psi-lambda-undirected} as,

\begin{eqnarray*}
\tilde{\Lambda}_{uv}(s) & = & \{(l,m):\,s_{l}=u,s_{m}=v,l,m\in1,\ldots,L^{(s)}\}\\
 & = & \{(l,m):\,s_{l}=u,s_{m}=v,l<m,l,m\in1,\ldots,L^{(s)}\}\\
 &  & +\{(l,m):\,s_{l}=u,s_{m}=v,l>m,l,m\in1,\ldots,L^{(s)}\}\\
 & = & \Lambda_{uv}(s)+\Lambda_{uv}^{T}(s)
\end{eqnarray*}

Therefore, the SGT 
for the bidirectional sequence in Eq.~\ref{eq:psi-main-len-sensi} can
be expressed as,


\begin{eqnarray*}
\tilde{\Psi}_{uv}(s) & = & \cfrac{\sum_{\forall(l,m)\in\tilde{\Lambda}_{uv}(s)}\phi_{\kappa}(d(l,m))}{|\tilde{\Lambda}_{uv}(s)|}\\
 & = & \cfrac{\sum_{\forall(l,m)\in\Lambda_{uv}(s)}\phi_{\kappa}(d(l,m))+\sum_{\forall(l,m)\in\Lambda_{uv}^{T}(s)}\phi_{\kappa}(d(l,m))}{|\Lambda_{uv}(s)|+|\Lambda_{uv}^{T}(s)|}\\
 & = & \cfrac{|\Lambda_{uv}(s)|\Psi_{uv}(s)+|\Lambda_{uv}^{T}(s)|\Psi_{uv}^{T}(s)}{|\Lambda_{uv}(s)|+|\Lambda_{uv}^{T}(s)|}
\end{eqnarray*}

Under the above assumption,

\begin{eqnarray}
\Lambda_{uv}(s) & \sim & \Lambda_{uv}^{T}(s)
 \label{eq:psi-lambda-similarity}
\end{eqnarray}

Therefore, the bidirectional SGT features can be approximated as

\begin{equation}
\tilde{\Psi}\sim\cfrac{\Psi+\Psi^{T}}{2}
\end{equation}

\subsection{SGT for Symbol Clustering}
\label{subsec:sgt-for-symbol-clustering}
\begin{figure}
\begin{raggedright}
\begin{minipage}[t]{0.45\columnwidth}%
\begin{center}
\subfloat[$u,v$ are closer than $u,w$. \label{fig:Illustrative-sequence-example-ele-clus}]{\begin{centering}
\includegraphics[scale=0.3]{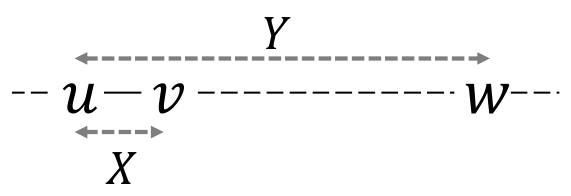}
\par\end{centering}
}
\par\end{center}%
\end{minipage}
\smallskip{}
\begin{minipage}[t]{0.45\columnwidth}%
\begin{center}
\subfloat[SGT's Graph view.]{\begin{centering}
\includegraphics[scale=0.22]{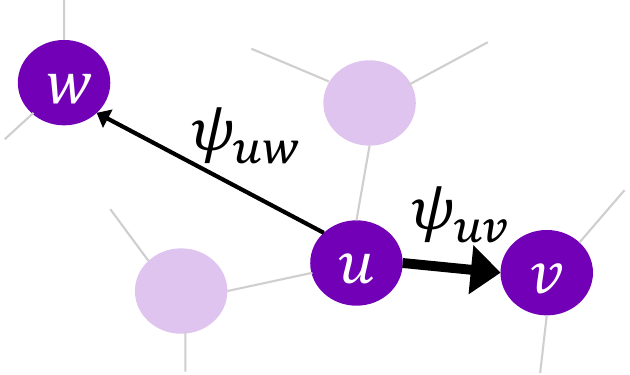}
\par\end{centering}
}
\par\end{center}%
\end{minipage}
\par\end{raggedright}
\caption{Illustrative example for symbol clustering.}
\end{figure}%

Node clustering in graphs is a classical problem solved by various
techniques, including spectral clustering, graph partitioning, and others.
SGT's graph interpretation facilitates grouping of symbols that
occur closely via any of these node clustering methods. 

This is because SGT gives larger weights to the
edges, $\psi_{uv}$, corresponding to symbol pairs that occur closely.
For instance, consider a sequence in Fig.~\ref{fig:Illustrative-sequence-example-ele-clus},
in which $v$ occurs closer to $u$ than $w$, also implying $E[X]<E[Y]$.
Therefore, in this sequence's SGT, the edge weight for
$u$$\rightarrow$$v$ should be greater than for $u$$\rightarrow$$w$,
i.e. $\psi_{uv}>\psi_{uw}$. 

From the assumption in \S\ref{sec:sgt-properties-short-long-term-features},
we will have, $E[|\Lambda_{uv}|]=E[|\Lambda_{uw}|]$. Therefore, $\psi_{uv}\propto E[\phi(X)]$
and $\psi_{uw}\propto E[\phi(Y)]$, and due to Condition~b on $\phi$
given in \S\ref{subsec:SGT-definition}, if $E[X]<E[Y]$, then $E[\psi_{uv}]>E[\psi_{uw}]$.

Moreover, for an effective clustering, it is important to bring the ``closer''
symbols in the sequence \emph{more} close in the graph space. In the
SGT's graph interpretation, it implies that $\psi_{uv}$
should go as high as possible to bring $v$ closer to $u$ in the
graph and vice versa for $(u,w)$. Thus, effectively, $\Delta=E[\psi_{uv}-\psi_{uw}]$
should be increased. It is proved in Appendix~B that $\Delta$ will increase with $\kappa$, if $\kappa d>1, \forall d $, where we have $d \in \mathbb{N}$. 

In effect, SGT enables the clustering of associated symbols. This
has real-world applications, such as finding the webpages (or products)
that are viewed (or bought) together.

\begin{figure}
\centering

  \includegraphics[scale=0.66]{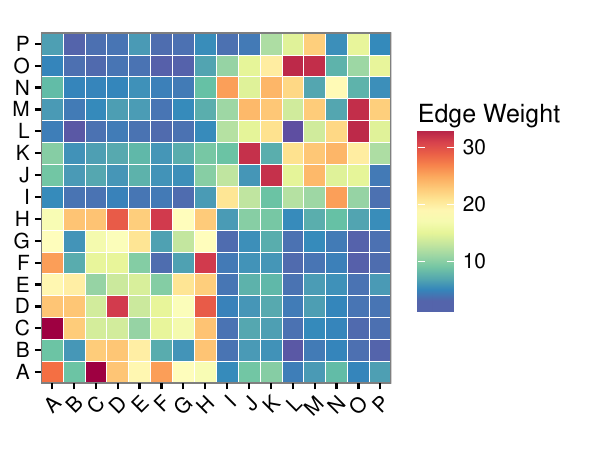}
  \caption{Node clustering experiment result.}
  \label{fig:Heat-map-showing-symbols'}

\end{figure}

\subsubsection{Validation}\label{sec:node-clustering-validation}
We validated the efficacy of the SGT extension given above (\S\ref{subsec:sgt-for-undirected-sequences}-\ref{subsec:sgt-for-symbol-clustering}) in another experiment presented here. Our main aim in this
validation is to perform symbol clustering assuming the sequences are bidirectional.
We set up a test experiment such that across different sequence clusters
some symbols occur closer to each other. We create a data set that
has sequences from three clusters and symbols belonging to two clusters
(symbols A-H in one cluster and I-P in another). The mean and standard deviation of the simulated sequence lengths is (103.9, 33.6). The noise is at 30-50\% and the number of underlying sequence clusters is equal to three.

This emulates a biclustering scenario in which sequences in different
clusters have distinct patterns; however, the pattern of closely
occurring symbols is common across all sequences. This is a complex
scenario in which clustering both sequences and symbols can be challenging.


Upon clustering the sequences, the F1-score is found to be 1.0. For
symbol clustering, we applied spectral clustering on the aggregated
SGT of all sequences, which yielded an accurate result with only
one symbol as mis-clustered. Moreover, a heat map in Fig.~\ref{fig:Heat-map-showing-symbols'}
clearly shows that symbols within the same underlying clusters have
significantly higher associations. Thus, it validates that SGT can
accurately cluster symbols along with clustering the sequences. 

\section{Conclusion}
\label{sec:Discussion-and-conclusion}

SGT is found to yield superior accuracy over other methods in sequence mining.
This can be attributed to SGT a) effectively capturing short-
and long-term patterns in both length- sensitive and insensitive problems, 
b) it inherently accounts for sequence mismatches to 
avoid false positives, and, c) it is robust to noise 
in sequence patterns. These attributes are discussed 
in detail in 
\S\ref{subsec:ability-to-work-in-length}-\ref{subsec:robust-to-noise}.
Moreover, SGT has significantly lower runtimes due to its computational efficiency and is also easy to implement. It can be further improved by implementing a sparse data structure for the $W$ matrices in the algorithms. In addition to the
above-mentioned applications, SGT can also be used for: a) element (symbol) clustering, b)
sequence database search, and c) sequence encoding, shown in the Applications section. 

Besides, some preliminary work shows the possibility
of a 2-D SGT applicable to image data allowing invariance to orientation. Moreover, other choices for the function $\phi$, such as Gaussian, or addition of a skip parameter $r$ (for addressing lag effects), for example, $e^{-\kappa \max{(d-r, 0)}}$, and application of concatenated (stacked) SGT features for different $\kappa$ may be taken as future
research. Furthermore, a formal approach for $\kappa$ selection can be developed.


\bibliographystyle{spmpsci}      
\bibliography{biblio}   



\appendix
\section*{Appendices}

\section{Mean and Variance of \large{$\psi_{uv}$}}
\label{appendix:theorem-mean-variance}

Consider an arbitrary sequence $s$, where the sequence has an ordered list of symbols. These symbols belong to a finite set $\mathcal{V}$. SGT embedding works by finding the dependencies between every pair of symbols $(u,v); u, v \in \mathcal{V}$.

To easily denote various $(u,v)$ pairs in $s$,
we use a term, $m^{th}$ neighboring pair, where an $m^{th}$
neighbor pair for $(u,v)$ will have $m-1$ other $u$'s in between.
A first neighbor is thus the immediate $(u,v)$ neighboring pairs,
while the $2^{nd}$-neighbor has one other $u$ in between, and so on (see
Fig.~\ref{fig:Representation-of-short-long} for illustration). The \emph{immediate
}neighbor mentioned in the assumption in Sec.~\ref{sec:sgt-properties-short-long-term-features}
is the same as the first neighbor defined here.

Based on the sequence patterns assumption in \S\ref{sec:sgt-properties-short-long-term-features} and
assuming $u,v$ occur uniformly in the sequence with probability $p$,
the expected number of first-neighbor $(u,v)$ pairs is given as $M=pL$.
Consequently, it is easy to show that the expected number of $m^{th}$
neighboring $(u,v)$ pairs is $(M-m+1)$, i.e., the second neighboring
$(u,v)$ pairs will be $(M-1)$, $(M-2)$ for the third, so on and
so forth, till one instance for the $M^{th}$ neighbor. 
The gap distance for an $m^{th}$ neighbor is given
as $Z_{1}=X;Z_{m}=X+\sum_{i=2}^{m}Y_{i},m=2,\ldots,M$.

Besides, the total number of $(u,v)$ pair instances will be $\sum_{m=1}^{M}m=\frac{M(M+1)}{2}$($=|\Lambda_{uv}|$,
by definition). Suppose we define a set that contains distances for
each possible $(u,v)$ pairs as $\mathcal{Z}=\{Z_{m}^{i},i=1,\ldots,(M-m+1);m=1,\ldots,M\}$.
Also, since $Z_{m}\sim N(\mu_{\alpha}+(m-1)\mu_{\beta},\sigma_{\alpha}^{2}+(m-1)\sigma_{\beta}^{2})$,
$\phi_{\kappa}(Z_{m})$ becomes a lognormal distribution. Thus,


\begin{eqnarray}
E[\phi_{\kappa}(Z_{m})] & = & e^{-\tilde{\mu}_{\alpha}-(m-1)\tilde{\mu}_{\beta}}\label{eq:Exp-phi-Z}\\
\textnormal{var}[\phi_{\kappa}(Z_{m})] & = & e^{-2\tilde{\mu}_{\alpha}'-2(m-1)\tilde{\mu}_{\beta}'}-e^{-2\tilde{\mu}_{\alpha}-2(m-1)\tilde{\mu}_{\beta}}\label{eq:Var-phi-Z}
\end{eqnarray}

where,


\begin{eqnarray}
\tilde{\mu}_{\alpha}=\kappa\mu_{\alpha}-\frac{\kappa^{2}}{2}\sigma_{\alpha}^{2} & ; & \tilde{\mu}'_{\alpha}=\kappa\mu_{\alpha}-\kappa^{2}\sigma_{\alpha}^{2}\nonumber \\
\tilde{\mu}_{\beta}=\kappa\mu_{\beta}-\frac{\kappa^{2}}{2}\sigma_{\beta}^{2} & ; & \tilde{\mu}'_{\beta}=\kappa\mu_{\beta}-\kappa^{2}\sigma_{\beta}^{2}\label{eq:mu-tildes}
\end{eqnarray}

Besides, the feature, $\psi_{uv}$, in Eq.~\ref{eq:psi-main-len-sensi}
can be expressed and further derived using \S\ref{subsec:arithmetico-geometric} as,


\begin{eqnarray}
E[\psi_{uv}] & = & \cfrac{\sum_{Z\in\mathcal{Z}}E[\phi_{\kappa}(Z)]}{M(M+1)/2}\nonumber \\
 & = & \cfrac{\sum_{m=1}^{M}(M-(m-1))e^{-\tilde{\mu}_{\alpha}-(m-1)\tilde{\mu}_{\beta}}}{M(M+1)/2}\nonumber \\
 & = & \cfrac{2}{pL+1}\cfrac{e^{-\tilde{\mu}_{\alpha}}}{\underbrace{\left|\left(1-e^{-\tilde{\mu}_{\beta}}\right)\left[1-\cfrac{1-e^{-pL\tilde{\mu}_{\beta}}}{pL(e^{\tilde{\mu}_{\beta}}-1)}\right]\right|}_{\gamma}}
\end{eqnarray}

This yields to the expectation expression in Eq.~\ref{eq:E-phi-AB-part-2-2}. Besides, the
variances will be


\begin{eqnarray*}
\textnormal{var}(\psi_{uv}) & = & \left(\cfrac{1}{pL(pL+1)/2}\right)^{2}\left[\left\{ \cfrac{e^{-2\tilde{\mu}'_{\alpha}}}{1-e^{-2\tilde{\mu}'_{\beta}}}\left(pL-\right.\right.\right.\\
 &  & \left.\left.e^{-2\tilde{\mu}'_{\beta}}\left(\cfrac{1-e^{-2pL\tilde{\mu}'_{\beta}}}{1-e^{-2\tilde{\mu}'_{\beta}}}\right)\right)\right\} -\\
 &  & \underbrace{\left.\left\{ \cfrac{e^{-2\tilde{\mu}{}_{\alpha}}}{1-e^{-2\tilde{\mu}{}_{\beta}}}\left(pL-e^{-2\tilde{\mu}_{\beta}}\left(\cfrac{1-e^{-2pL\tilde{\mu}{}_{\beta}}}{1-e^{-2\tilde{\mu}{}_{\beta}}}\right)\right)\right\} \right]}_{\pi}
\end{eqnarray*}


\begin{eqnarray*}
\textnormal{var}(\psi_{uv}) & = & \begin{cases}
\left(\cfrac{1}{pL(pL+1)/2}\right)^{2}\pi & ;\text{length sensitive}\\
\left(\cfrac{1}{p(pL+1)/2}\right)^{2}\pi & ;\text{length insensitive}
\end{cases}
\end{eqnarray*}

\subsection{Arithmetico-Geometric Series}
\label{subsec:arithmetico-geometric}

The sum of a series, where the $k^{th}$ term for $k\geq1$ can be
expressed as,


\begin{eqnarray*}
t_{k} & = & \left(a+(k-1)d\right)br^{k-1}
\end{eqnarray*}

is called an arithmetico-geometric because of a combination of
arithmetic series term $(a+(k-1)d)$, where $a$ is the initial term
and common difference $d$, and geometric $br^{k-1}$, where $b$
is the initial value and common ratio being $r$.

Suppose the sum of the series till $n$ terms is denoted as,


\begin{eqnarray}
S_{n} & = & \sum_{k=1}^{n}\left(a+(k-1)d\right)br^{k-1}\label{eq:appendix-A-1}
\end{eqnarray}

Without loss of generality we can assume $b=1$ for deriving the expression
for $S_{n}$ (the sum for any other value of $b$ can be easily obtained
by multiplying the expression for $S_{n}$ with $b$). Expanding Eq.~\ref{eq:appendix-A-1},


\begin{eqnarray}
S_{n} & = & a+(a+d)r+\ldots+(a+(n-1)d)r^{n-1}\label{eq:Appendix-A-2}
\end{eqnarray}

Now multiplying $S_{n}$ with $r$,


\begin{eqnarray}
rS_{n} & = & ar+(a+d)r^{2}+\ldots+(a+(n-1)d)r^{n}\label{eq:Appendix-A-3}
\end{eqnarray}

Subtracting Eq.~\ref{eq:Appendix-A-3} from \ref{eq:Appendix-A-2},
if $|r|<1$, else we subtract the latter from the former, we get,


\begin{eqnarray*}
\left|(1-r)S_{n}\right| & = & \left|\left[a+(\cancel{a}+d)r+\ldots+(\cancel{a}+\cancel{(n-1)}d)r^{n-1}\right]\right.\\
 &  & \left.-\left[\cancel{ar}+(\cancel{a}+\cancel{d})r^{2}+\ldots+(a+(n-1)d)r^{n}\right]\right|\\
 & = & \left|a+d(r+r^{2}+\ldots+r^{n-1})-(a+(n-1)d)r^{n}\right|\\
 & = & \left|a+\cfrac{dr(1-r^{n-1})}{1-r}-\left(a+(n-1)d\right)r^{n}\right|.
\end{eqnarray*}

Therefore,


\begin{eqnarray}
S_{n} & = & \left|\cfrac{1}{1-r}\left[a+\cfrac{dr(1-r^{n-1})}{1-r}-\left(a+(n-1)d\right)r^{n}\right]\right|\label{eq:Appendix-A-4}
\end{eqnarray}

or, for any value of $b$,


\begin{eqnarray}
S_{n} & = & b\left|\cfrac{1}{1-r}\left[a+\cfrac{dr(1-r^{n-1})}{1-r}-\left(a+(n-1)d\right)r^{n}\right]\right|\label{eq:Appendix-A-5}
\end{eqnarray}

\section{$\mathbf{W}^{(\kappa)}$ independent with respect to $\kappa$}
\label{appendix:wk-independence-proof}

\textit{Proof:}

We have $\mathbf{W}^{(\kappa)} = [W^{(\kappa)}_{uv}],\,u,v\in \mathcal{V}$ where 

\begin{equation}
\label{eq:proof-wk-independence-1}
    W^{(\kappa)}_{uv} = \sum_{\forall (l,m) \in \Lambda_{uv}(s)}e^{-\kappa|m-l|}    
\end{equation}{}

To prove the independence of $\mathbf{W}^{(\kappa)}$ with respect to $\kappa$, we need to show $\mathbf{w}_{u\cdot}^{(\kappa)} \indep  \mathbf{w}_{u\cdot}^{(\kappa + \delta)}$ where $\mathbf{w}_{u\cdot}$ is a column in $\mathbf{W}$.

Without loss of generality assume $\Lambda_{uv}(s) = 1$ and replacing $|m-l|$ with $x, x>0$ in Equation~\ref{eq:proof-wk-independence-1} the column $\mathbf{w}_{u\cdot}$ for $\kappa$ and $\kappa + \delta$ will be, $\mathbf{w}_{u\cdot}^{(\kappa)} = [e^{-\kappa x_i}]$ and $\mathbf{w}_{u\cdot}^{(\kappa + \delta)} = [e^{-\kappa x_i}]$ where $i = 1, \ldots, |\mathcal{V}|$.

$\mathbf{w}_{u\cdot}^{(\kappa)}$ and $\mathbf{w}_{u\cdot}^{(\kappa+\delta)}$ will be dependent iff there is a nontrivial solution for $a,b$ in Equation~\ref{eq:proof-wk-independence-2}.

\begin{equation}
\label{eq:proof-wk-independence-2}
    a\mathbf{w}_{u\cdot}^{(\kappa)} - b\mathbf{w}_{u\cdot}^{(\kappa + \delta)} = 0
\end{equation}{}

Solving this for $a$ and $b$ by taking any element $i$ in $\mathbf{w}_{u\cdot}$.

\begin{eqnarray*}
&a w_{ui}^{(\kappa)} - b w_{ui}^{(\kappa+\delta)}&=0\\
\implies&a e^{-\kappa x_i} - b e^{-(\kappa+\delta) x_i}&=0\\
\implies& e^{-\kappa x_i}(a - b e^{-\delta x_i})&=0\\
\end{eqnarray*}

Since $e^{-\kappa x_i}\neq 0$, we have $a = b e^{-\delta x_i}$. Plugging this in the equation for another element $j$ in $\mathbf{w}_{u\cdot}$ we get,

\begin{eqnarray*}
&a w_{uj}^{(\kappa)} - b w_{uj}^{(\kappa+\delta)}&=0\\
\implies&b e^{-\delta x_i} e^{-\kappa x_j} - b e^{-(\kappa+\delta) x_j}&=0\\
\implies& b e^{-\kappa x_j}(e^{-\delta x_i} - e^{-\delta x_j})&=0\\
\end{eqnarray*}

Since $e^{-\delta x_i} \neq e^{-\delta x_j} \forall i,j$, and $e^{-\kappa x_j} \neq 0$ we have $b=0$. Consequently, $a=0$.

Therefore, Equation~\ref{eq:proof-wk-independence-2} has a trivial solution $a,b=0$. Thus, $\mathbf{w}_{u\cdot}^{(\kappa)}$ is independent with respect to $\kappa$ implying independence of $\mathbf{W}^{(\kappa)}$.

\section{Proof for Symbol Clustering}
We have, $\frac{\partial\Delta}{\partial\kappa}=\cfrac{\partial}{\partial\kappa}E[\phi_{\kappa}(X)-\phi_{\kappa}(Y)]=E[\frac{\partial}{\partial\kappa}\phi_{\kappa}(X)-\frac{\partial}{\partial\kappa}\phi_{\kappa}(Y)]$.
For $E[X]<E[Y]$, we want, $\frac{\partial\Delta}{\partial\kappa}>0$,
in turn, $\frac{\partial}{\partial\kappa}\phi_{\kappa}(X)>\cfrac{\partial}{\partial\kappa}\phi_{\kappa}(Y)$.
This will hold if $\frac{\partial^{2}}{\partial d\partial\kappa}\phi_{\kappa}(d)>0$,
that is, slope, $\frac{\partial}{\partial\kappa}\phi_{\kappa}(d)$
is increasing with $d$. For an exponential expression for $\phi$
(Eq.~\ref{eq:phi-expression}), the above condition holds true if
$\kappa d>1$. Hence, under these conditions, the \textit{separation}
increases as we increase the tuning parameter, $\kappa$.

\section{Sequence Simulation}
\label{sec:sequence-simulation}

In this section, we explain the sequence generation for
Exp~1-2 in \S~\ref{sec:Experimental-Analysis}.

\begin{figure}
\begin{centering}
\includegraphics[scale=0.26]{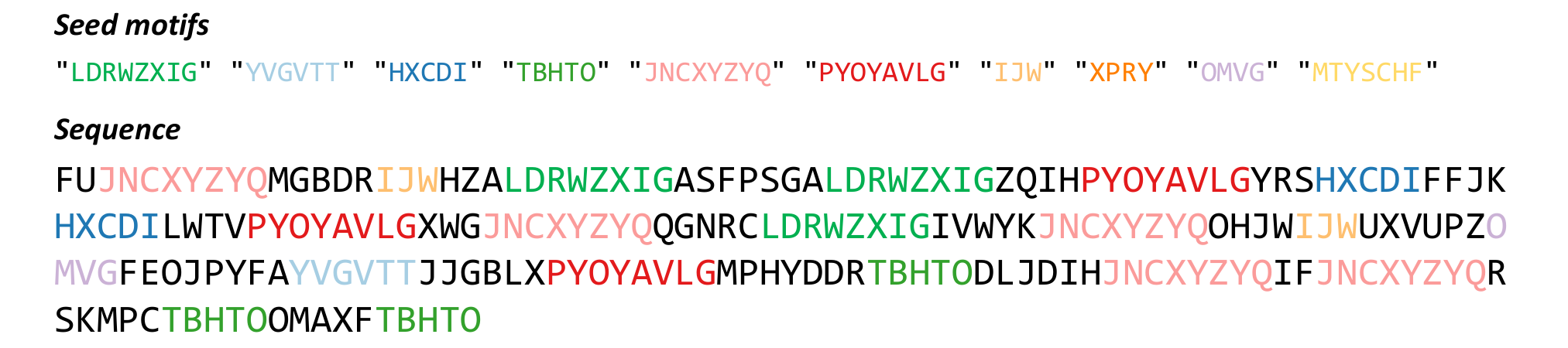}
\par\end{centering}
\caption{A simulated sequence from seed \textit{motifs}.\label{fig:sample-simulated-sequence}}
\end{figure}

A sequence is generated by randomly selecting a string from the motif set and placing them in random order between arbitrary 
strings. These interspersed arbitrary symbols are the noise in a sequence.
Fig.~\ref{fig:sample-simulated-sequence} shows an example of
a sequence generated from a set of seed motifs. About 50\% of the sequence is noise---arbitrary strings. Note that due to the
random motif selection, a simulated sequence does not necessarily
contain all seed motifs.

\subsection*{Generating sequence clusters}

Suppose we have to generate $K$ sequence clusters. We first
randomly simulate $K$ sets of motifs. In each set, the motifs
are of random lengths (between 2-8 in our simulations). The
size of a set is also randomly chosen (between 6-11 in this paper).
Sequences are then generated from each motif set as described above.

\subsection*{What are overlapping clusters?}

\begin{figure}
\begin{centering}
\includegraphics[scale=0.26]{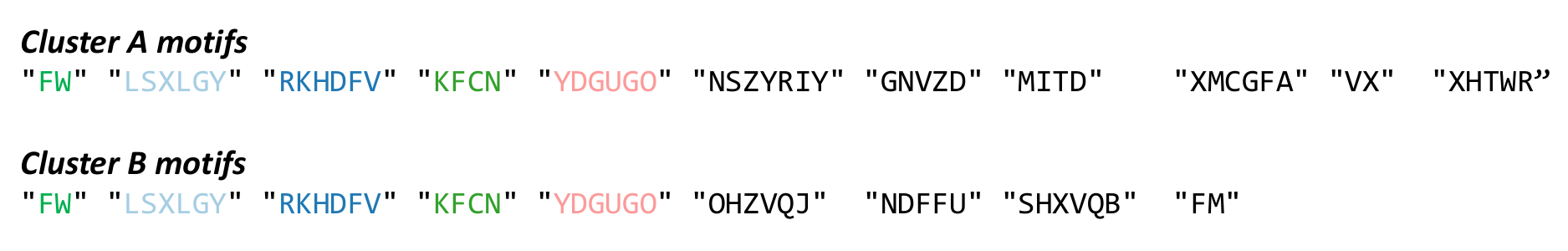}
\par\end{centering}
\caption{Motif sets of overlapping clusters.\label{fig:overlapping-clusters}}
\end{figure}%

In our experiments, we have to test the efficacy of clustering methods
when the clusters are difficult to separate. This is the case when
the clusters \textit{overlap}. In traditional multidimensional data
in Euclidean space, it means the centroids of the clusters
are close. In our problem, overlapping clusters imply that they 
have some common seed motifs. In other words, the intersection
of the clusters' motif sets is not null. Thus, 0\% overlap implies
the intersection of motif sets is null, and a 100\% overlap implies
the intersection is equal to the union. Fig.~\ref{fig:overlapping-clusters} shows an example of overlapping motif sets of two
clusters.

\section{Code Repository and Data Sets}
\label{sec:code-repository-and-data-sets}

The source code and data sets are available on GitHub as \url{https://github.com/cran2367/sgt}. Its python package is also provided at \url{https://pypi.org/project/sgt/}.

The python package can be installed as follows.

\begin{lstlisting}
$ pip install sgt
\end{lstlisting}{}

\end{document}